\pdfoutput=1

\documentclass[lettersize,journal]{IEEEtran}
\usepackage{amsmath,amsfonts}
\usepackage{xfrac}
\usepackage{algorithmic}
\usepackage{algorithm}
\usepackage{array}
\usepackage{booktabs}
\usepackage[caption=false,font=normalsize,labelfont=sf,textfont=sf]{subfig}
\usepackage{textcomp}
\usepackage{stfloats}
\usepackage{url}
\usepackage{arydshln}
\usepackage{verbatim}
\usepackage{epsfig}
\usepackage{graphicx}
\usepackage{threeparttable}
\usepackage{multicol}
\usepackage{multirow}
\usepackage{diagbox}
\usepackage{cite}
\def\etal{\emph{et al. }}

\begin{document}

\title{Scale-Net: Learning to Reduce Scale Differences for Large-Scale Invariant Image Matching}

\author{Yujie Fu, Yihong Wu
\thanks{This paper was produced by the IEEE Publication Technology Group. They are in Piscataway, NJ.}
\thanks{Manuscript received}}

%

\markboth{None}
{Shell \MakeLowercase{\textit{et al.}}: A Sample Article Using IEEEtran.cls for IEEE Journals}

\IEEEpubid{0000--0000/00\$00.00~\copyright~2021 IEEE}

\maketitle

\begin{abstract}
Most image matching methods perform poorly when encountering large scale changes in images.
To solve this problem, firstly, we propose a scale-difference-aware image matching method (SDAIM) that reduces image scale differences before local feature extraction, via resizing both images of an image pair according to an estimated scale ratio.
Secondly, in order to accurately estimate the scale ratio, we propose a covisibility-attention-reinforced matching module (CVARM) and then design a novel neural network, termed as Scale-Net, based on CVARM. 
The proposed CVARM can lay more stress on covisible areas within the image pair and suppress the distraction from those areas visible in only one image.
Quantitative and qualitative experiments confirm that the proposed Scale-Net has higher scale ratio estimation accuracy and much better generalization ability compared with all the existing scale ratio estimation methods.
Further experiments on image matching and relative pose estimation tasks demonstrate that our SDAIM and Scale-Net are able to greatly boost the performance of representative local features and state-of-the-art local feature matching methods.
\end{abstract}


\section{Introduction}

\IEEEPARstart{E}{stablishing} pixel-level correspondences between two images is an essential basis for a wide range of computer vision tasks such as structure from motion \cite{7780814, schnberger2016pixelwise}, augmented reality \cite{8887213}, visual localization \cite{9229078}, and Simultaneous localization and mapping (SLAM) \cite{campos2020orbslam3}. Such correspondences are usually estimated by sparse local feature extraction and matching \cite{lowe2004distinctive, 6126544, yi2016lift, ono2018lf, detone2018superpoint, 8953622, r2d2, luo2020aslfeat, cavalli2020handcrafted, 9310246, sarlin2020superglue}. 
A local feature consists of a keypoint and a descriptor. 
But the scale invariance of both existing keypoint detectors and descriptors is not enough to deal with large scale changes \cite{zhou2017progressive}. 
Few inlier correspondences can be established by matching local features under the circumstances of large scale changes in images, which is called as the scale problem of local features in this paper. 
If the scale difference between two images is small, we call that the two images are at related scale levels in scale space \cite{lindeberg1998feature, lindeberg2013scale}.

\begin{figure}[t]
	\centering
	\begin{center}
		\includegraphics[width=0.9\linewidth]{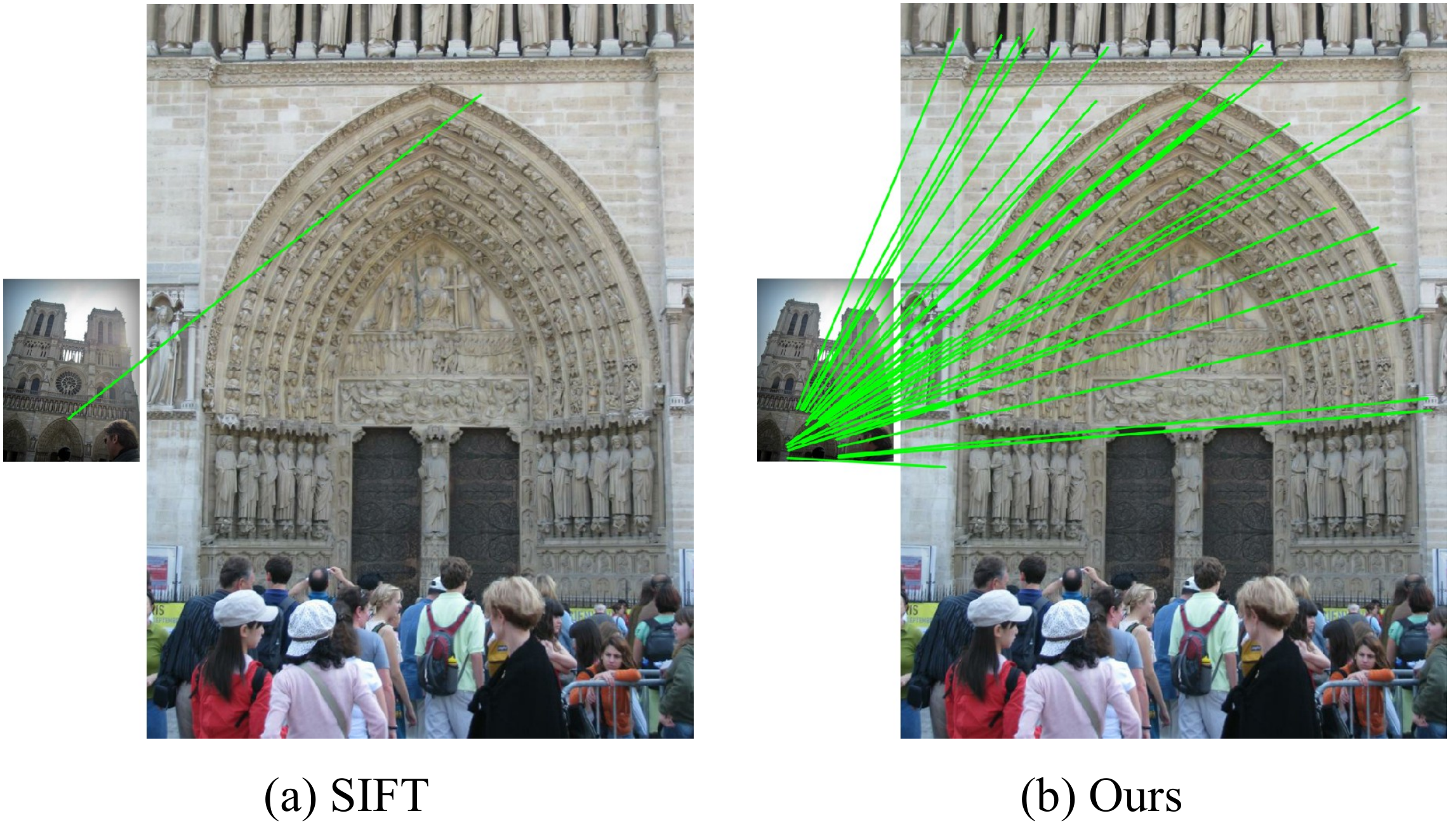}
	\end{center}
	\caption{2k SIFT keypoints are extracted from a challenging image pair with large scale change. Correspondences of SIFT (a) and our method (b) after RANSAC verification are displayed. 
		Only those correspondences conforming to the ground truth epipolar geometry are drawn.}
	\label{fig:cover}
\end{figure}

To alleviate the scale problem of local features, the multi-scale feature extraction method based on the image pyramid (MSFE-IP) is widely used \cite{lowe2004distinctive, bay2006surf, 8953622, luo2020aslfeat}. Given an image, local features are extracted from several neighbouring scale levels of original image scale level. MSFE-IP improves the robustness of local features to scale changes.
However, if the scale difference is too large, MSFE-IP still only establishes very few inlier correspondences, as depicted in Figure \ref{fig:cover}(a). 
The reasons are as follows. 
As observed by the literature \cite{jegou2008hamming, li2015pairwise, zhou2017progressive}, given an inlier correspondence, it is very likely that the two local features linked by this correspondence are extracted from two related scale levels. If a correspondence consists of two local features extracted from two unrelated scale levels, it is very likely that this correspondence is wrong.
Given an image pair to be matched, the scale difference of the image pair is unknown. MSFE-IP only samples several scale levels near original image scale levels. Thus, MSFE-IP is not able to ensure that there exist related scale levels among the sampled ones, especially when the scale difference is very large.
When the scale difference between two images is not very large, most of the scale levels sampled by MSFE-IP are related so that many inlier correspondences can be established. 
However, if the scale difference is too large, few or even no sampled scale levels are related so that few inlier correspondences will be established.\IEEEpubidadjcol

In a word, the scale problem of local features can be ascribed to too few related scale levels to a large extent. 
To deal with this problem, we propose a scale-difference-aware image matching approach (SDAIM), as shown in Figure \ref{fig:SDAIM}(a). Given an image pair to be matched, firstly, its scale difference is estimated. Then both images are resized according to the estimated scale difference so that the scale difference of the image pair is greatly reduced before multi-scale local feature extraction.
This approach guarantees that most sampled scale levels are related and restricts most local features to be extracted from the related scale levels. 
As shown in Figure \ref{fig:cover}(b), after enhanced by our SDAIM, SIFT can establish much more inlier correspondences.
The challenge is how to estimate the scale difference or a scale ratio (Section \ref{sec_definitions})? 

So far, image scale ratio estimation problem has drawn little attention.
Zhou \etal used a Bag-of-Features (BoF) model to encode the image pyramids of an image pair and exhaustively matched the representation vectors of sampled scale levels to estimate the scale ratio \cite{zhou2017progressive}.
A box embedding for images is proposed to estimate the scale ratio by Rau \etal \cite{rau2020predicting}. 
But the generalization ability of both the above methods is weak, which results from the loss of visual overlap information during image encoding processes. To make the most of visual overlap information, we propose a covisibility-attention-reinforced matching module (CVARM). 
Given two images and their dense feature maps computed by a convolutional neural network (CNN), it takes image dense feature maps as input and exhaustively matches image patch pairs. 
It is able to lay more stress on covisible areas and suppress those areas visible in only one image. 
Because the dense feature maps computed by a vanilla CNN is sensitive to scale changes, we design a multi-scale feature extraction and fusion module to compute image dense feature maps. 
We also design a scale ratio regressor module which follows CVARM and estimates the scale ratio. 
The above three modules are chained together and form a neural network, termed as Scale-Net, whose accuracy and generalization ability are both experimentally proved to be much better than those of the methods proposed by Zhou \etal \cite{zhou2017progressive} and Rau \etal \cite{rau2020predicting}. 
Although there does not exist a large-scale training set with scale ratio annotation, we prove that Scale-Net can be trained by synthetic data in an end-to-end manner conveniently. 
Besides, we create a benchmark dataset containing large scale changes of images based on IMC-PT dataset \cite{jin2020image}. 
We call this dataset as Scale-IMC-PT dataset. It can be used to evaluate the performance of local features on relative pose estimation task with large scale changes in outdoor scenes. It can also be used to finetune Scale-Net.

In summary, our contributions are listed as follows:
\begin{enumerate}
	\item{We analyze the reasons for the scale problem of local features. And we propose a scale-difference-aware image matching approach, termed as SDAIM, that has an intrinsic invariance to scale changes.}
	\item{We present a novel neural network, termed as Scale-Net, based on our proposed covisibility-attention mechanism, which can emphasize the visual overlaps of an image pair and accurately estimate the scale ratio of the image pair. 
	Scale-Net can be easily integrated into the proposed SDAIM.}
	\item{Extensive experiments confirm that SDAIM can remarkably enhance the performance of local features under large image scale changes and that Scale-Net has good generalization ability and high scale ratio estimation accuracy.}
\end{enumerate}

\section{Related Work}
Since the purpose of this paper is to mitigate the scale problem of local features, below we briefly review local feature extraction and local feature matching in the literature. Because image scale ratio estimation is an essential step in our method, we also review some scale ratio estimation methods.

\subsection{Local Feature Extraction}

Local feature extraction consists of keypoint detection and feature description. 
Various handcrafted keypoint detectors have been proposed over the past few decades. FAST \cite{rosten2006machine} and Harris \cite{harris1988combined} are pioneering corner detectors. 
But neither of them is invariant to large scale changes. A simple way to cope with scale changes is to extract keypoints at several scales. Mikolajczyk and Schmid used the extrema over scale of the Laplacian-of-Gaussian to select the scale of interest points picked by multi-scale Harris detector \cite{mikolajczyk2004scale}. SIFT uses Difference-of-Gaussian operator to select blobs over multiple scale levels \cite{lowe2004distinctive}. KAZE applies Hessian detector to a nonlinear diffusion scale space \cite{alcantarilla2012kaze}. In recent years, several learning-based detectors have been proposed \cite{verdie2015tilde, savinov2017quad, barroso2019key}. TILDE focuses on devising a keypoint detector which is robust to imaging changes of weather and lighting conditions \cite{verdie2015tilde}. Savinov \etal proposed a transformation-invariant keypoint detector that can be trained in an unsupervised manner \cite{savinov2017quad}.
Key.Net is one of the few learned detectors taking the scale problem into account, which uses a multi-scale neural network to detect robust features \cite{barroso2019key}. 
Given keypoints, high level information can be captured in patches around keypoints by descriptors. Handcrafted descriptors encode image patches based on grayscale and gradient information  \cite{lowe2004distinctive, calonder2010brief}. Learning-based descriptors usually use convolutional neural networks to encode image patches so that they are able to dig out higher level information than handcrafted counterparts. And the convolutional neural networks often trained by discriminability-oriented loss functions \cite{tian2017l2, mishchuk2017working, tian2019sosnet, liu2019gift, luo2018geodesc, luo2019contextdesc}. But few of descriptors focus on scale invariance. GIFT uses group convolutions to fuse features extracted from the transformed versions of an image to obtain a descriptor which is invariant to scale changes \cite{liu2019gift}. There are also several detect-and-describe methods \cite{r2d2, 8953622, luo2020aslfeat, detone2018superpoint, ono2018lf, shen2019rf}. ASLFeat \cite{luo2020aslfeat}, LF-Net \cite{ono2018lf} and RF-Net \cite{shen2019rf} adopt different strategies to deal with scale changes. 
ASLFeat utilizes the inherent pyramidal features of a CNN \cite{luo2020aslfeat}. LF-Net resizes the feature maps computed by a ResNet \cite{he2016deep} several times to simulate several scale levels \cite{ono2018lf}. RF-Net treats different layers of a CNN as different scale levels \cite{shen2019rf}. 
But all the above methods only take several scale levels near original image scale levels into account.
When facing large scale changes, most sampled scale levels of the above methods are unrelated, which results in their degradation. To solve this problem, our method reduces image scale differences before sampling scale levels so that most sampled scale levels are related.

\subsection{Local Feature Matching}

The most widely used matching procedure contains the following steps: nearest neighbour (NN) matching, mutual NN checking,  ratio test \cite{lowe2004distinctive} and RANSAC verification \cite{fischler1981random, raguram2012usac, barath2019magsac, barath2018graph}. 
In recently years, several new matching methods have been proposed.
Given putative correspondences of feature points in two views, 
PointCN uses a deep neural network to tell inlier correspondences from the outlier ones \cite{yi2018learning}. OANet introduces both local and global context of sparse correspondences into learned outlier rejection methods \cite{9310246}. 
AdaLAM is a handcrafted method, which utilizes spatial consistency in a hierarchical pipeline for effective outlier rejection \cite{cavalli2020handcrafted}.
But PointCN, OANet and AdaLAM are all outlier rejection methods. When facing large scale changes, there are very few inlier correspondences among putative correspondences computed by NN matching. In that case, under the help of PointCN, OANet and AdaALAM, the number of inlier correspondences is still very small.
SuperGlue is a ground-breaking matching method. 
It uses an attention-based context aggregation mechanism to refine descriptors and  find correspondences \cite{sarlin2020superglue}. However, under large scale changes, due to the low repeatability of keypoint detectors, SuperGlue is not able to establish many inlier correspondences, either.
In summary, under large scale changes, because local features are extracted from mismatched scale levels, existing matching methods can not help local features out of trouble. In order to make sure that most local features are extracted from related scale levels, the proposed SDAIM narrows scale differences before local feature extraction. Therefore, SDAIM can significantly boost the performance of local feature matching methods when facing large scale differences.

\subsection{Image Scale Ratio Estimation}

There are very few methods of estimating the scale ratio of an image pair.
Rau \etal introduced an image box embedding to estimate the image scale ratio \cite{rau2020predicting}. But the model of Rau \etal is scene-specific. An image box embedding model can only make sense in the training scene. 
Zhou \etal used a scale level matching method based on Bag-of-Features model (SLM-BoF) to determine the scale ratio of an image pair for better scale-invariant image matching \cite{zhou2017progressive}. 
By contrast, we propose a learning-based method to estimate the scale ratio, which has much better generalization ability than the image box embedding model \cite{rau2020predicting} and much higher accuracy than SLM-BoF \cite{zhou2017progressive}.

\section{Deep Scale-Difference-Aware Image Matching}

\subsection{Definitions}
\label{sec_definitions}
Our goal is narrowing the scale difference between two images before feature extraction. But how to determine that there is no scale difference? We adopt the same definition proposed by Rau \etal \cite{rau2020predicting}:

There is no scale difference between two images when visual overlaps occupy approximately the same number of pixels in each of the two images. Visual overlaps are image areas picturing the same 3D object surfaces, which are marked by red rectangles in Figure \ref{fig:SDAIM}.

We also make a definition for the image scale ratio:
If the scale ratio between image $I_1$ and image $I_2$ is $s$, $I_1$ should be resized to $s$ times its original size so that there is nearly no scale difference between the resized $I_1$ and $I_2$. The scale ratio definition is denoted by $\phi(\cdot,\cdot)$, as shown in:

\begin{equation}
\phi\left(I_1, I_2\right)=s, \phi\left(I_2, I_1\right)=\frac{1}{s}
\label{def_sr}
\end{equation}
Note that the scale ratio definition is asymmetric.

\subsection{Scale-Difference-Aware Image Matching}
\label{sec_SDAIM}

\begin{figure*}[ht]
	\begin{center}
		\includegraphics[width = 0.9\linewidth]{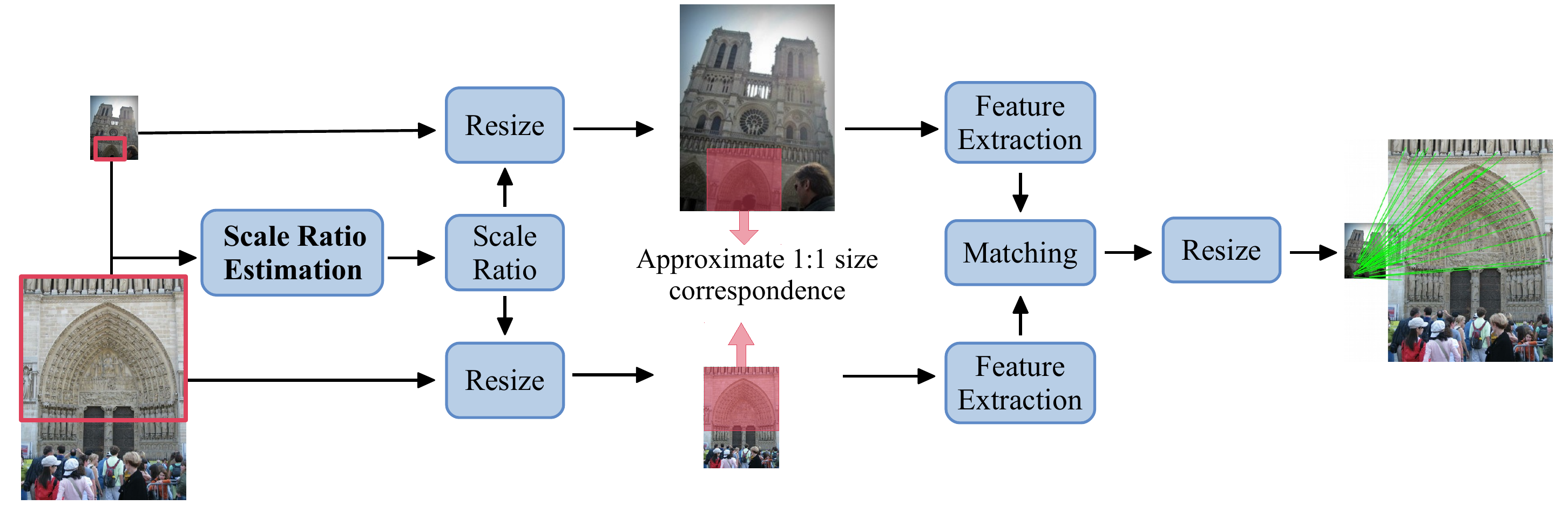}
	\end{center}
	\caption{\textbf{Scale-Difference-Aware Image Matching}: Firstly, the scale ratio of two views is estimated. Then, both images are resized according to the scale ratio. The following steps are feature extraction and matching. The last step is restoring feature locations according to the original image sizes. Visual overlaps are approximately marked by red rectangles.
		\textbf{Best viewed in color.}}
	\label{fig:SDAIM}
\end{figure*}

Traditional multi-scale feature extraction methods based on the image pyramid (MSFE-IP) ignore the scale difference between two images and only extract local features from the scale levels near original image scale levels \cite{lowe2004distinctive, bay2006surf, 8953622, luo2020aslfeat}. 
By contrast, we propose a scale-difference-aware image matching (SDAIM) approach that reduces the scale difference before local feature extraction, as illustrated in Figure \ref{fig:SDAIM}.
Given two images to be matched, firstly, a scale ratio is estimated.
Secondly, both images are resized according to the scale ratio. The resizing process is illustrated as below.
 
Given two images, $I_1$ and $I_2$,  assume that $ \phi\left(I_1, I_2\right)=s$. 
$I_1$ is resized to $ s^{-0.5} $ times its original size. $I_2$ is resized to $ s^{0.5} $ times its original size.
Then local features are extracted from the resized image pair, which is followed by feature matching. Finally, spatial locations of the local features extracted from the resized images are restored according to the sizes of original images.

\begin{figure}[t]
	\begin{center}
		\includegraphics[width=0.9\linewidth]{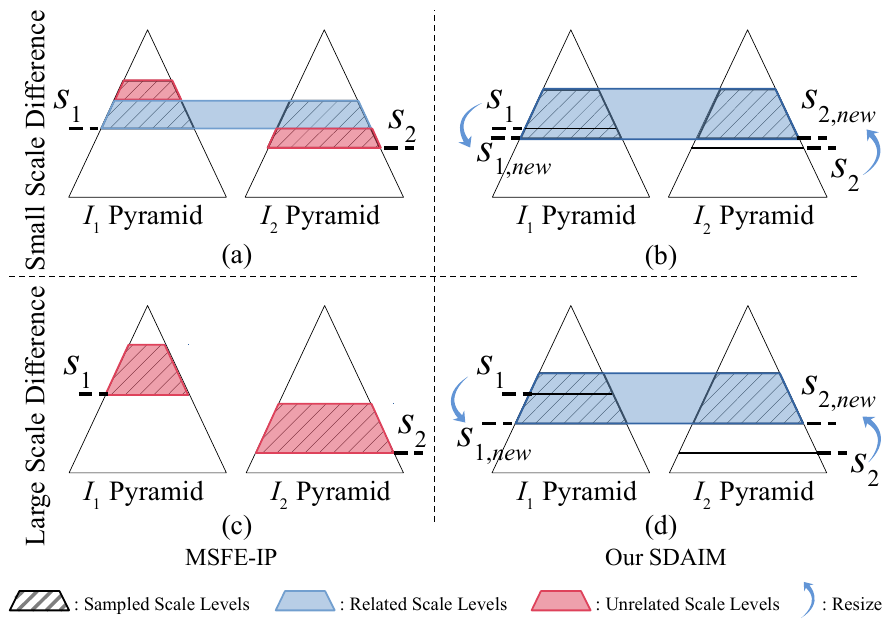}
	\end{center}
	\caption{\textbf{Comparison between the multi-scale feature extraction (MSFE-IP) and our approach.} $s_1$ represents the scale level of image $I_1$. $s_2$ represents the scale level of image $I_2$. $s_{1, new}$ represents the scale level of the resized $I_1$. $s_{2, new}$ represents the scale level of the resized $I_2$. \textbf{Best viewed in color.}}
	\label{fig:pyr}
\end{figure}

The advantages of the proposed SDAIM over MSFE-IP are depicted in Figure \ref{fig:pyr}. 
There are four subfigures. 
In each subfigure, a white triangle represents the Gaussian Pyramid of an image. 
If there is no scale difference between image $I_1$ and image $I_2$, their scale levels, $s_1$ and $s_2$, are at the same height. In that case, $s_1$ and $s_2$ are related scale levels. 
Without loss of generality, we assume that images are only down-sampled in MSFE-IP.
In MSFE-IP, as shown in Figure \ref{fig:pyr}(a) and (c), only neighbouring scale levels of original image scale levels are sampled. Local features are only extracted at those sampled scale levels. If the scale difference between $I_1$ and $I_2$ is not very large, as shown in Figure \ref{fig:pyr}(a), some sampled scale levels of $I_1$ are related with sampled scale levels of $I_2$. The related scale levels are marked in blue. Many inlier matches can be established between these related scale levels. But there are some unrelated scale levels which are marked in red. Local features extracted at these unrelated scale levels increase the probability of mismatch. 
What's worse, when the scale difference is large, all the sampled scale levels are unrelated, as shown in Figure \ref{fig:pyr}(c). 
In that case, very few matches are right, as vividly depicted in Figure \ref{fig:cover}(a). 

Our approach is capable of solving the above problems. As shown in Figure \ref{fig:pyr}(b) and (d), no matter how large the scale difference is, the scale difference between resized $I_1$ and resized $I_2$ is greatly reduced before feature extraction so that almost all sampled scale levels of $I_1$ are related with those of $I_2$. In other words, our approach can select better scale levels for local feature matching than MSFE-IP, which results in sufficient inlier matches for downstream tasks. The superiority of our approach is significant. As shown in Figure \ref{fig:cover}(b), after enhanced by our approach, SIFT can establish much more inlier correspondences.

\subsection{Scale Ratio Estimation Network}

\begin{figure*}[ht]
	\begin{center}
		\includegraphics[width = 0.95\linewidth]{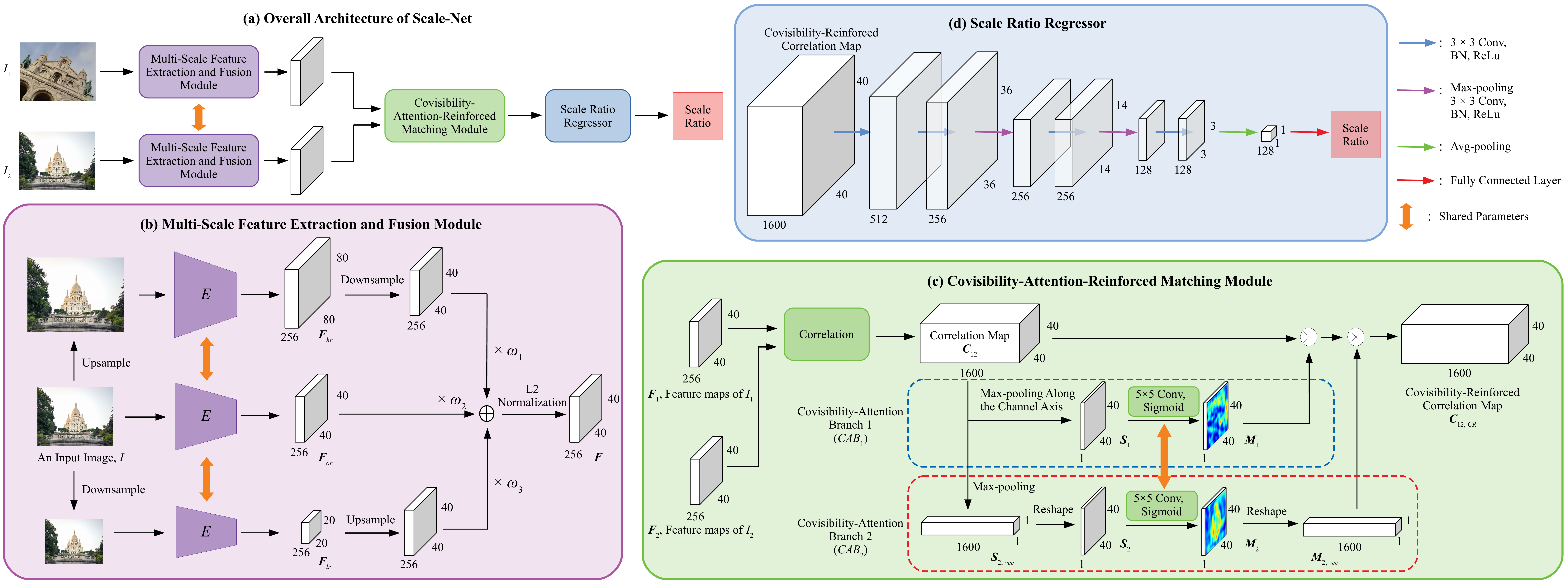}
	\end{center}
	\caption{
		\textbf{(a) Overall Architecture of Scale-Net.}
		\textbf{(b) Multi-Scale Feature Extraction and Fusion Module.} This module uses a vanilla CNN to extract dense feature maps from three scale levels of an image and outputs the L2-normalized weighted sum of the dense feature maps.
		\textbf{(c) Covisibility-Attention-Reinforced Matching Module.} This module takes dense feature maps of an image pair as input, exhaustively matches the patches of the image pair and obtains a global cost volume. Then it max-pools the global cost volume in one branch and max-pools the global cost volume along the channel axis in another branch to compute the covisibiity attention maps. At last, it multiplies the global cost volume by the covisibiity attention maps to emphasize covisible areas and suppresses the distraction from non-covisible areas.
		\textbf{(d) Scale Ratio Regressor.} This module takes a cost volume as input and outputs the scale ratio of the input image pair.
		\textbf{Best viewed in color with 300\% zoom in.}
	}
	\label{fig:scale-net}
\end{figure*}

The above SDAIM needs a scale ratio estimation method.
A neural network, termed as Scale-Net, is proposed to estimate the scale ratio of a given image pair. Its overall architecture is shown in Figure \ref{fig:scale-net}(a). It consists of three components: Multi-Scale Feature Extraction and Fusion Module (MSFEF), Covisibility-Attention-Reinforced Matching Module (CVARM) and Scale Ratio Regressor. 
Given two images, $ I_{1} $ and $ I_{2} $, firstly, MSFEF is used to extract dense feature maps of $ I_{1} $ and $ I_{2} $. The L2-normalized dense feature maps of $ I_{1} $ and $ I_{2} $ are denoted by $ \textit{\textbf{F}}_1 $ and $ \textit{\textbf{F}}_2 $.
Secondly, CVARM takes $ \textit{\textbf{F}}_1 $ and $ \textit{\textbf{F}}_2 $ as input and outputs a correlation map. 
At last, Scale Ratio Regressor takes the correlation map as input and calculates the scale ratio of the image pair. 

\textbf{Multi-Scale Feature Extraction and Fusion Module (MSFEF).} This module is devised to extract multi-scale dense feature maps from an image, as shown in Figure \ref{fig:scale-net}(b).
In this module, the input image is downsampled once and upsampled once. And a three-level Gaussian Pyramid is obtained. A CNN denoted by $ \textit{E}\left(\cdot\right) $ is used to extract dense feature maps from each level of the pyramid. Then, the dense feature maps extracted from the higher-resolution image $ \textit{\textbf{F}}_{hr} $, is downsampled. The dense feature maps extracted from the lower-resoluion image $ \textit{\textbf{F}}_{lr} $, is upsampled. 
The sizes of downsampled $ \textit{\textbf{F}}_{hr} $ and upsampled $ \textit{\textbf{F}}_{lr} $ are identical with that of the dense feature maps extracted from the original image $ \textit{\textbf{F}}_{or} $. The final output of this module is the L2-normalized weighted sum of $ \textit{\textbf{F}}_{hr} $, $ \textit{\textbf{F}}_{or} $ and $ \textit{\textbf{F}}_{lr} $. This module is summarized as below:
\begin{equation}\label{MSFEF}
\begin{aligned}
\textit{\textbf{F}}_{sum} = \ 
& \omega_{1}\left(Ds\left(E\left(Us\left(I\right)\right)\right)\right) \\
& +\omega_{2}\left(E\left(I\right)\right) + \omega_{3}\left(Us\left(E\left(Ds\left(I\right)\right)\right)\right)
\end{aligned}
\end{equation}

\begin{equation}
\textit{\textbf{F}} = L2Norm\left(\textit{\textbf{F}}_{sum}\right)
\end{equation}
where $ DS\left(\cdot\right) $ represents downsampling, $ Us\left(\cdot\right) $ represents upsampling, $ \omega_{1} $, $ \omega_{2} $ and $ \omega_{3} $ are weights.

\textbf{Covisibility-Attention-Reinforced Matching Module (CVARM).} According to the scale ratio definition (Section \ref{sec_definitions}), the scale ratio depends on the visual overlaps between $ I_{1} $ and $ I_{2} $. Therefore, Scale-Net should pay much attention to the information of covisible areas for accurate scale ratio estimation. To this end, Covisibility-Attention-Reinforced Matching Module (CVARM) is proposed.
The diagram of this module is shown in Figure \ref{fig:scale-net}(c). 
Given the dense feature maps extracted by MSFEF, $ \textit{\textbf{F}}_1, \textit{\textbf{F}}_2\in\mathbb{R}^{h\times w\times c} $, there are $ \left(h\times w\right) \times \left(h\times w\right) $ pairs of local feature descriptors between $ \textit{\textbf{F}}_1 $ and $ \textit{\textbf{F}}_2 $ in total. The scalar products of all these descriptor pairs are computed and form a correlation map $ \textit{\textbf{C}}_{12}\in \mathbb{R}^{h\times w\times\left(h\times w\right)}$, which is identical with the matching layer proposed by Rocco \etal \cite{8434328}. The calculation process of $ \textit{\textbf{C}}_{12} $ is shown in:

\begin{equation}\label{correlationMap}
\textit{\textbf{C}}_{12}\left(i, j, k\right) = \textit{\textbf{F}}_1\left(i, j\right)^{T}\textit{\textbf{F}}_2\left(i_{k}, j_{k}\right)
\end{equation}

\noindent where $ k=h\left(i_{k}-1\right)+j_{k} $, $ i\in \left[1, h\right], i_{k}\in \left[1, h\right] $, $ j\in \left[1, w\right], j_{k}\in \left[1, w\right] $.
Each \textit{c}-dimensional vector of $ \textit{\textbf{F}}_1 $ or $ \textit{\textbf{F}}_2 $ is a descriptor of a patch in image $ I_{1} $ or image $ I_{2} $.
Therefore, the raw correlation map, $ \textit{\textbf{C}}_{12} $, is constructed by exhaustively matching patches of $ I_{1} $ and $ I_{2} $. 
$ \textit{\textbf{C}}_{12}\left(i, j, k\right) $ is just the inner product or the cosine similarity between patch $ \left(i, j\right) $ in $ I_{1} $ and patch $ \left(i_{k}, j_{k}\right) $ in $ I_{2} $. 
Thus, $ \textit{\textbf{C}}_{12} $ contains much information of non-covisible areas. The information of non-covisible areas may have a bad influence on scale ratio estimation. To deal with this problem, we improve the matching layer.
Our new matching layer can lay more stress on the covisible areas and suppress the disturbance resulting from non-covisible areas. Our new matching layer is detailed as below.

In the remainder of this article, each element of $ \textit{\textbf{C}}_{12} $ is called as the similarity between two patches.
Assume a certain patch \textit{A} is in $ I_{1} $ and a certain patch \textit{B} is in $ I_{2} $.
After $ \textit{\textbf{C}}_{12} $ is obtained, patch \textit{A} has been matched with all the patches in $ I_{2} $. $ \textit{\textbf{C}}_{12} $ contains $ \left(h\times w\right) $ similarity values related to \textit{A}, each of which is the similarity between patch \textit{A} and a certain patch in $ I_{2} $.
Generally, if patch \textit{A} and patch \textit{B} picture the same 3D surface, their similarity is high. Furthermore, if patch \textit{A} belongs to the covisible areas, the highest similarity value among all the  $ \left(h\times w\right) $ similarity values related to \textit{A} is large. Thus, the highest similarity value of a certain patch (HSVP) can be used to determine whether this patch belongs to covisible areas.

Based on the above analysis, we devise two covisibility-attention branches to dig out covisible areas in $ I_{1} $ and $ I_{2} $ from the raw correlation map $ \textit{\textbf{C}}_{12} $. The two covisibility-attention branches are called $ {CAB}_{1} $ and $ {CAB}_{2} $ in Figure \ref{fig:scale-net}(c). 
The corresponding patch in $ I_1 $ of descriptor $ \textit{\textbf{F}}_1\left(i, j\right) $ is denoted by $ P_{1, ij} $.
In $ \textit{\textbf{C}}_{12} $, the $ \left(h\times w\right) $-dimensional vector $ \textit{\textbf{C}}_{12}\left(i, j\right) $ stores all the similarity values related to $ P_{1, ij} $. In $ {CAB}_{1} $, $ \textit{\textbf{C}}_{12} $ is max-pooled along the channel axis. Thereafter, the HSVPs of all patches in $ I_{1} $ are obtained and form a similarity map $ \textit{\textbf{S}}_{1}\in \mathbb{R}^{h\times w\times 1} $. Then, a convolution operation with the filter size of $ 5\times 5 $, $ f^{5\times 5} $, and a sigmoid function, $ \sigma $, are used to refine $ \textit{\textbf{S}}_{1} $. Finally, a covisibility score map $ \textit{\textbf{M}}_{1}\in \mathbb{R}^{h\times w\times 1}$ is obtained. $ \textit{\textbf{M}}_{1}\left(i, j\right) $ represents the probability of belonging to covisible areas of $ P_{1, ij} $. In another word, $ \textit{\textbf{M}}_{1} $ is a soft mask that indicates covisible areas in image $ I_{1} $.
The calculation process of $ \textit{\textbf{M}}_{1} $ is summarized as below:

\begin{equation}\label{covisibilityMap1}
\textit{\textbf{M}}_{1}=\sigma\left(f^{5\times 5}\left(MaxPool_{Channel}\left(\textit{\textbf{C}}_{12}\right)\right)\right)
\end{equation}

The corresponding patch in $ I_{2} $ of descriptor $ \textit{\textbf{F}}_2\left(i_{k}, j_{k}\right) $ is denoted by $ P_{2, i_{k}j_{k}} $.
The \textit{k}-th channel of $ \textit{\textbf{C}}_{12} $ is a matrix denoted by $ \textit{\textbf{T}}_{k} $. $ \textit{\textbf{T}}_{k} $ stores all the similarity values related to $ P_{2, i_{k}j_{k}} $. In $ {CAB}_{2} $, $ \textit{\textbf{C}}_{12} $ is max-pooled globally. Thereafter, the HSVPs of all patches in $ I_{2} $ are obtained and form a $ \left(h\times w\right) $-dimensional vector $ \textit{\textbf{S}}_{2, vec} $. 
According to Eq. (\ref{correlationMap}), $ \textit{\textbf{S}}_{2, vec} $ can be reshaped to a similarity map $ \textit{\textbf{S}}_{2}\in \mathbb{R}^{h\times w \times 1} $. This reshaping operation is denoted by $ Rsp_{1}\left(\cdot\right) $, which is detailed in Eq. (\ref{reshape1}).
The following step is using a convolution operation with the filter size of $ 5\times 5 $, $ f^{5\times 5} $, and a sigmoid function, $ \sigma $, to refine $ \textit{\textbf{S}}_{2} $. The refined $ \textit{\textbf{S}}_{2} $ is called as $ \textit{\textbf{M}}_{2} $, which is a soft mask that indicates covisible areas in image $ I_{2} $.
At last, $ \textit{\textbf{M}}_{2} $ is reshaped to $ \textit{\textbf{M}}_{2, vec}\in \mathbb{R}^{1\times 1\times \left(h\times w\right)} $. This reshaping operation is denoted by $ Rsp_{2}\left(\cdot\right) $, which is detailed in Eq. (\ref{reshape2}).
The above calculation process is summarized as below:

\begin{equation}\label{reshape1}
Rsp_1\left(\cdot\right): \textit{\textbf{S}}_{2}\left(i_{k}, j_{k}\right)\leftarrow \textit{\textbf{S}}_{2, vec}\left(k\right), k=h\left(i_{k}-1\right)+j_{k}
\end{equation}
\begin{equation}\label{reshape2}
Rsp_2\left(\cdot\right): \textit{\textbf{M}}_{2, vec}\left(k\right)\leftarrow \textit{\textbf{M}}_{2}\left(i_{k}, j_{k}\right), k=h\left(i_{k}-1\right)+j_{k}
\end{equation}
\begin{equation}\label{covisibilityMap2}
\textit{\textbf{M}}_{2, vec} = Rsp_{2}\left(\sigma\left(f^{5\times 5}\left(Rsp_{1}\left(MaxPool\left(\textit{\textbf{C}}_{12}\right)\right)\right)\right)\right)
\end{equation}

$ \textit{\textbf{M}}_{2, vec} $ and $ \textit{\textbf{M}}_{1} $ are used to reduce the information of non-covisible areas and emphasize covisibility information in $ \textit{\textbf{C}}_{12} $, as shown in:
\begin{equation}\label{attentionApply}
\textit{\textbf{C}}_{12, CR} = \textit{\textbf{M}}_{2, vec}\otimes \left(\textit{\textbf{M}}_{1}\otimes \textit{\textbf{C}}_{12}\right)
\end{equation}
where $ \otimes $ denotes element-wise multiplication. During multiplication, $ \textit{\textbf{M}}_{1} $ is broadcasted along the channel axis, $ \textit{\textbf{M}}_{2, vec} $ is broadcasted along the spatial dimension.
Then, a covisibility-reinforced correlation map $ \textit{\textbf{C}}_{12, CR} $ is obtained. 
In the end, \textbf{Scale Ratio Regressor} takes $ \textit{\textbf{C}}_{12, CR} $ as input and ouputs a scale ratio between $ I_{1} $ and $ I_{2} $, which is shown in Figure \ref{fig:scale-net}(d).

\textbf{Dual Consistent Loss.} $D\equiv\left\lbrace \left( I_{i1}, I_{i2}, s_i \right) \right\rbrace_i^N$ represents a training set. $I_{i1}$ and $I_{i2}$ form an image pair, whose ground truth scale ratio is $s_i$. Scale-Net is denoted by $ SN $. Assume that the scale ratio between $I_{i1}$ and $I_{i2}$ estimated by Scale-Net is $ \hat{s}_{i} $, which is denoted by $ SN\left(I_{i1}, I_{i2}\right) = \hat{s}_{i} $. 
Given a training sample $ \left( I_{i1}, I_{i2}, s_i \right) $, there is a dual training sample $ \left( I_{i2}, I_{i1}, s_i^{-1} \right) $ according to Eq. (\ref{def_sr}).
Assume that $ SN\left(I_{i2}, I_{i1}\right) = \hat{s}^{\prime}_{i} $. Then, we propose a dual loss $ l_{d} $, as shown in Eq. (\ref{eq_dual}). Besides, $ SN\left(I_{i1}, I_{i2}\right) $ and $ SN\left(I_{i2}, I_{i1}\right) $ are expected to be consistent with each other according to Eq. (\ref{def_sr}). Thus, we devise a consistent loss $ l_{c} $, as shown in Eq. (\ref{eq_cons}).

\begin{equation}\label{eq_dual}
l_{d} = \frac{1}{2N}\sum_{i=1}^{N}\left[\left(\log_2{\frac{\hat{s}_{i}}{s_{i}}}\right)^2+\left(\log_2{\hat{s}^{\prime}_{i}s_{i}}\right)^2\right]
\end{equation}

\begin{equation}\label{eq_cons}
l_{c} = \frac{1}{N}\sum_{i=1}^{N}\left(\log_2{\hat{s}_{i}} + \log_2{\hat{s}^{\prime}_{i}}\right)^2
\end{equation}

$ l_{d} $ and $ l_{c} $ are combined to train our Scale-Net, as shown in:
\begin{equation}\label{eq_final_loss}
L = \lambda_{d}l_{d} + \lambda_{c}l_{c}
\end{equation}
where $ \lambda_{d} $ and $ \lambda_{c} $ are weights.

\section{Experiments}

\subsection{Datasets Containing Large Scale Differences}

To our knowledge, there does not exist a dataset containing large scale differences, whose samples are sufficient to train a neural network. 
Thus, we create a large dataset based on MS-COCO 2014 dataset \cite{lin2014microsoft} by synthetically generating image pairs with scale differences. We call this dataset as Scale-COCO dataset. We use Scale-COCO dataset to train our Scale-Net.
But Scale-COCO dataset does not contain self-occlusion, which is an important challenge in correspondence estimation. Therefore, we create another dataset based on IMC-PT dataset \cite{jin2020image}. We call this dataset as Scale-IMC-PT dataset. 
ES-HP dataset, proposed by Liu \etal \cite{liu2019gift}, contains considerable image scale differences. We resort to ES-HP dataset and Scale-IMC-PT dataset to demonstrate the effect of Scale-Net and SDAIM. 
The above datasets are detailed as below.

\begin{figure}[htbp]
	\centering
	\includegraphics[width=0.9\linewidth]{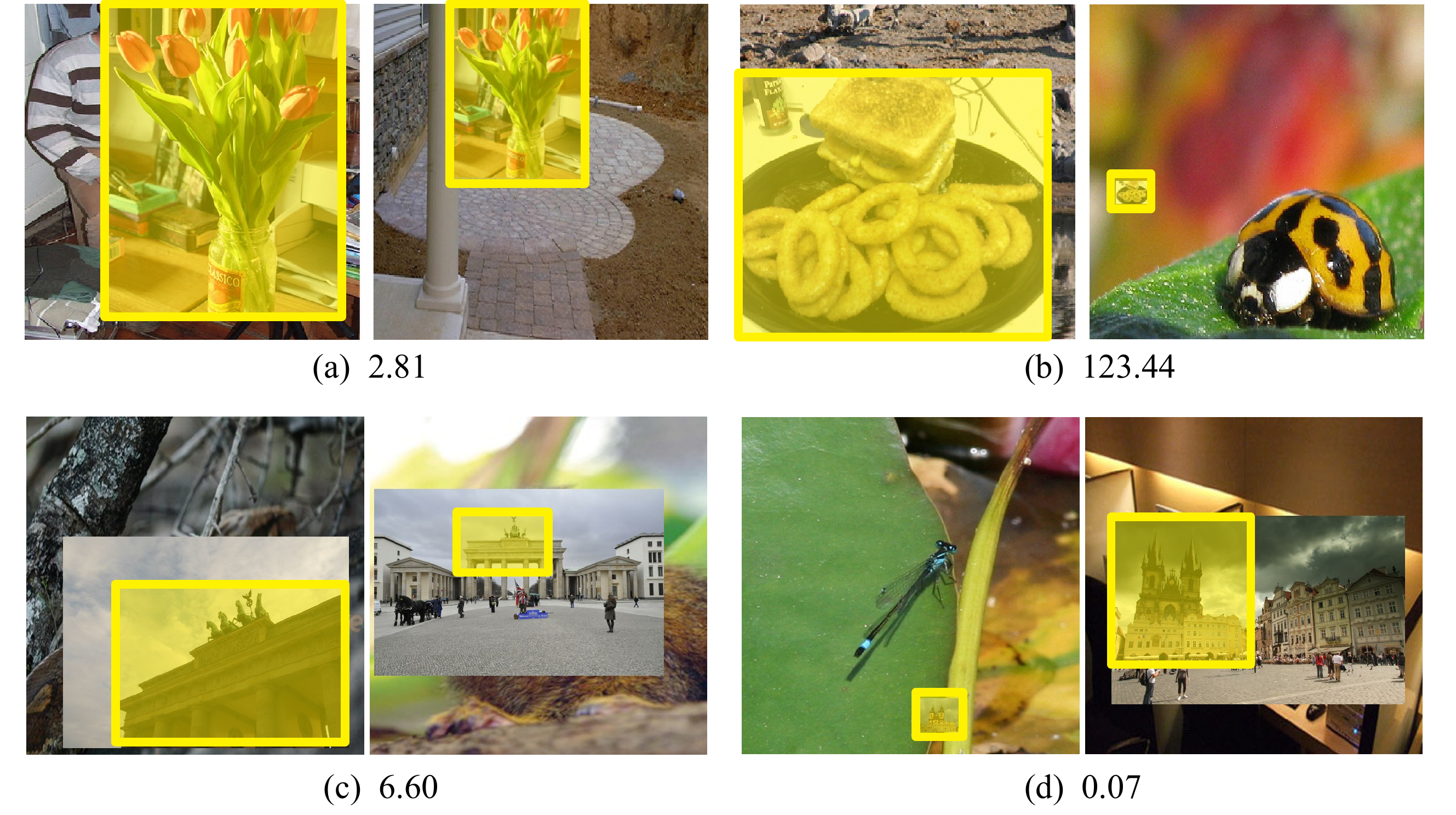}
	\caption{\textbf{Training image pairs.} (a) and (b) are the samples of Scale-COCO dataset. (c) and (d) are the samples of Scale-IMC-PT dataset. Visual overlaps are roughly marked by yellow rectangles. The number below each image pair is the ground truth scale ratio of the image pair. \textbf{Best viewed in color.}}
	\label{fig:traindata}
\end{figure}

\textbf{Scale-COCO.} We randomly select a real number, $ m\in\left[0, 7\right] $, and get a scale ratio $ s=2^m $.
Two different images, $ I_{b1} $ and $ I_{b2} $ are selected as background images from the training set of ImageNet ILSVRC dataset \cite{russakovsky2015imagenet}.
Training image pairs are generated by downsampling or upsampling. 
To generate a training sample by downsampling, an image $ I $ is randomly selected from MS-COCO 2014 training set. $ I $ is downsampled to $ 2^{-m} $ times its original size. The downsampled $ I $ is denoted by $ I^\prime $. $ I $ and $ I^\prime $ are pasted into $ I_{b1} $ and $ I_{b2} $ respectively. Then a training image pair is obtained, whose visual overlaps are $ I $ and $ I^\prime $.
To generate a training sample by upsampling, the image $ I $ from MS-COCO 2014 training set is upsampled to $ 2^{m} $ times its original size. Thereafter, $ I^\prime $ is generated by cropping the upsampled $ I $ at the center. Then we paste $ I $ and $ I^\prime $ into $ I_{b1} $ and $ I_{b2} $ respectively and obtain a training image pair.
Half of the training samples are generated by downsampling. The other half are generated by upsampling. There are 165,566 training image pairs in total.
Two training image pairs are shown in Figure \ref{fig:traindata}(a) and (b).
The scale ratio range of Scale-COCO dataset is $ \left[0.0078, 1\right)\cup \left(1, 128\right] $.

\textbf{Scale-IMC-PT.} There are 15 outdoor scenes with image poses, camera intrinsic parameters and semi-dense depth maps in IMC-PT dataset \cite{jin2020image}. But ground truth scale ratios are not provided in IMC-PT dataset.
The scale ratio of an image pair depends heavily on the visual overlaps. And visual overlaps can be found accurately by means of the method proposed by Rau \etal \cite{rau2020predicting} when image poses, camera intrinsic parameters and semi-dense depth maps are known. 
Thus, we make use of the method proposed by Rau \etal to generate image pairs with ground truth scale ratios. The procedure of ground truth scale ratio annotation is detailed as below.

\begin{figure}[tbph]
	\centering
	\includegraphics[width=0.6\linewidth]{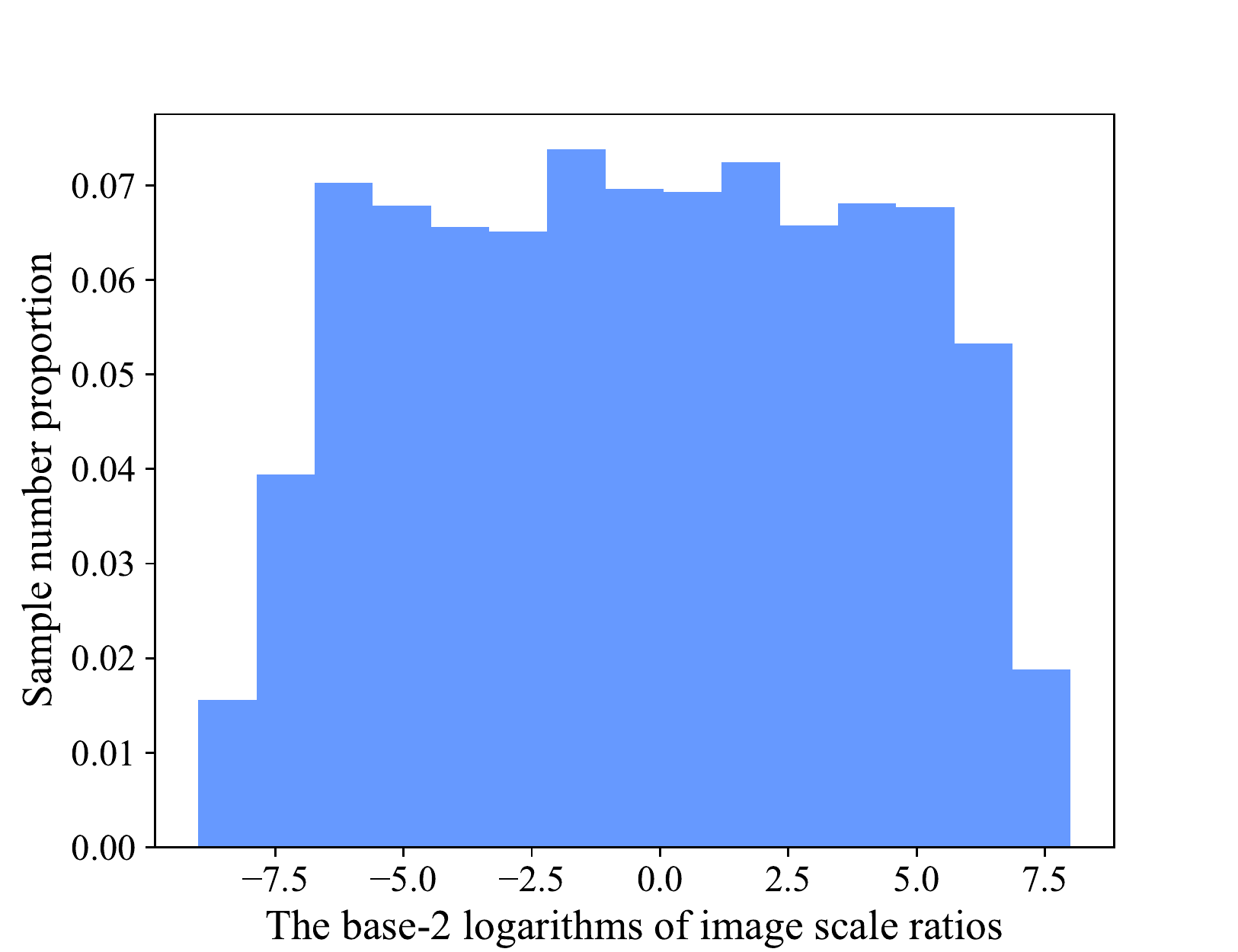}
	\caption[scale ratio dist]{The image scale ratio histograms of Scale-IMC-PT dataset.}
	\label{fig:srdist}
\end{figure}

Assume that there are visual overlaps between image $ I_{1} $ and image $ I_{2} $.
Firstly, visible point clouds of $ I_{1} $ and $ I_{2} $ in the world coordinate system can be easily obtained using their poses, their camera intrinsic parameters and their semi-dense depth maps. Visible point clouds of $ I_{1} $ and $ I_{2} $ are denoted by $ P_{1} $ and $ P_{2} $, respectively.
Secondly, given a 3D point $ p $ in $ P_{1} $, we compute the Euclidean distances between $ p $ and all 3D points in $ P_{2} $ and call the shortest one as the distance between $ p $ and $ P_{2} $. If the distance between $ p $ and $ P_{2} $ is smaller than a threshold $ \tau $, $ p $ is visible in $ I_{2} $. Then, in $ P_{1} $, we compute the number of 3D points that are visible in $ I_{2} $. This number is called as $ V_{1} $. In $ P_{2} $, we also compute the number of 3D points that are visible in $ I_{1} $. This number is called as $ V_{2} $.
At last, the scale ratio between $ I_{1} $ and $ I_{2} $ is $ \sfrac{V_{1}}{V_{2}} $, i.e., $ \phi\left(I_1, I_2\right)=\sfrac{V_{1}}{V_{2}} $, where $ \phi\left(\cdot\right) $ has the same meaning as the one in Eq. (\ref{def_sr}).
But the ground truth scale ratios of most of the image pairs in IMC-PT dataset are smaller than 4 or larger than 0.25, which are far smaller than the scale changes that we encounter in practical applications. Thus, we resize original image pairs in IMC-PT dataset to enlarge scale changes. Because our Scale-Net needs two images with same sizes, resized image pairs are pasted into background images.
Two annotation results are shown in Figure \ref{fig:traindata}(c) and (d).
The scale ratio range of Scale-IMC-PT dataset is $ \left[0.002, 1\right)\cup \left(1, 512\right] $.
The image scale ratio distribution histogram of Scale-IMC-PT dataset is shown in Figure \ref{fig:srdist}. 

Our Scale-IMC-PT dataset is divided into a training set and a test set.
The image pairs in test set are not pasted into background images.
There are 446685 image pairs in the training set. There are 39778 image pairs in the test set.
Training image pairs are selected from the following scenes: Brandenburg Gate, Buckingham Palace, Colosseum Exterior, Grand Place Brussels, Pantheon Exterior, Prague Old Town Square, Reichstag, Taj Mahal, Temple Nara Japan, Westminster Abbey, Trevi Fountain.
Test image pairs are selected from the following scenes: Notre Dame Front Facade (NDFF), Palace of Westminster (PW), Sacre Coeur (SC), St. Peter’s Square (SPS).

\textbf{Scale-MegaDepth.} MegaDepth is a large-scale outdoor dataset \cite{li2018megadepth}. 
Rau \etal arranged four outdoor scenes based on MegaDepth dataset \cite{rau2020predicting}. The four scenes are Notre-Dame (ND), Big Ben (BB), Venice (Ve) and Florence (Fl) respectively.
In each scene, the ground truth scale ratios of image pairs can be easily computed by the method proposed by Rau \etal \cite{rau2020predicting}.
We call this dataset including four outdoor scenes as Scale-MegaDepth dataset. Its scale ratio range is $ \left[0.002, 1\right)\cup\left(1, 512\right] $.


\textbf{ES-HP.} ES-HP dataset \cite{liu2019gift} is based on HPatches dataset \cite{balntas2017hpatches}. It is created by artificially scaling the images in HPatches dataset. Its scale ratio range is $ \left[0.0625, 0.125\right] \cup \left[8, 16\right] $.

\subsection{Implementation Details}
The ResNet-18 architecture \cite{he2016deep}, pretrained on ImageNet \cite{deng2009imagenet} and truncated after the Conv4\_2 layer, is used as the feature extractor $ E\left(\cdot\right) $, in MSFEF. In $ E\left(\cdot\right) $, only those parameters of Conv4\_2 layer are fine-tuned during training. We use Scale-COCO dataset to train Scale-Net. The Adam optimizer is used to train Scale-Net for 30 epochs. The base learning rate is set to 0.0003. Our implementation is made in PyTorch with one TITAN RTX cards. The mini-batch size is 48. The training images are resized to $ 640\times 640 $. 
We apply small random perspective transformations to training images in order to simulate viewpoint changes in real scenes.
In Eq. (\ref{eq_final_loss}), we set $ \lambda_{d}=1 $, $ \lambda_{c}=1 $.
When evaluating on the test set of Scale-IMC-PT dataset and Scale-MegaDepth dataset, we further finetune Scale-Net on the training set of Scale-IMC-PT dataset for 1 epoch.

\subsection{Image Scale Ratio Estimation}
\label{sec_sre}
\begin{figure}[htbp]
	\centering
	\includegraphics[width=0.9\linewidth]{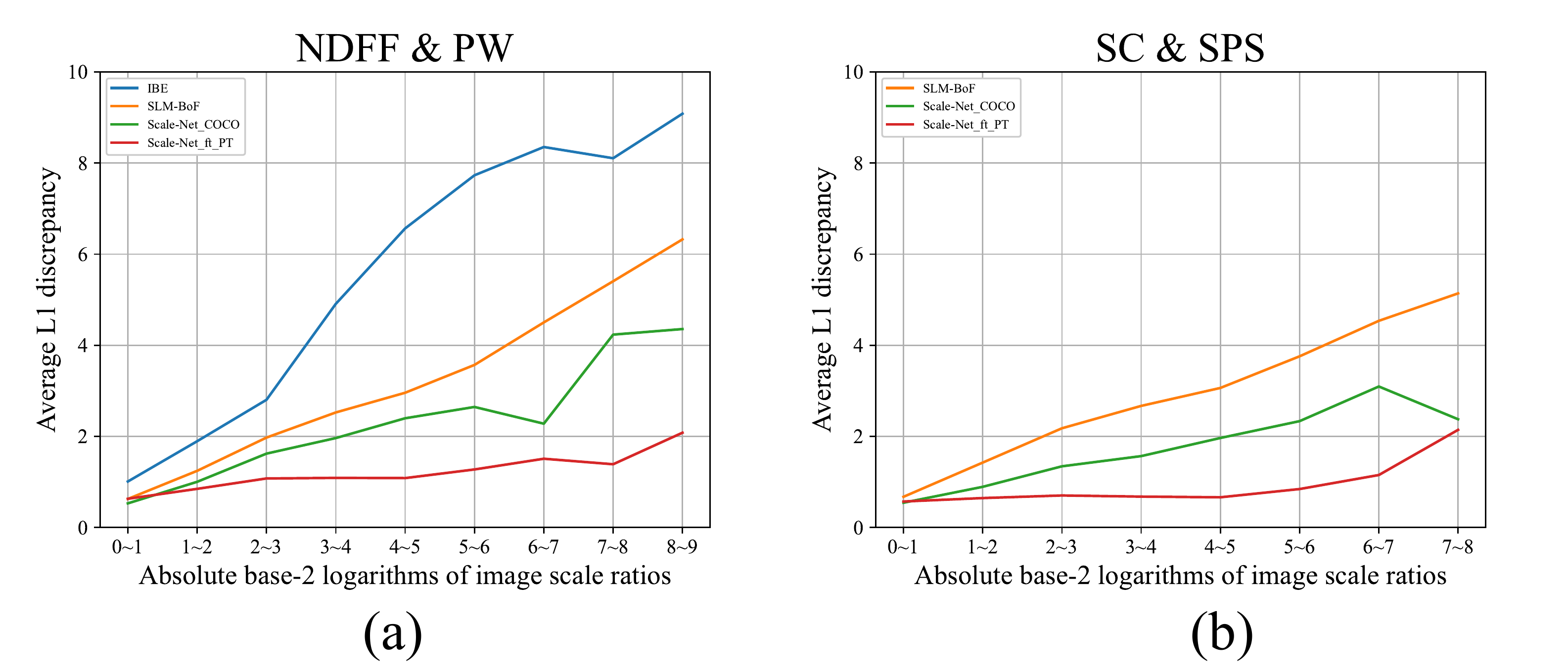}
	\caption{\textbf{The line graph of scale ratio estimation error over scale differences of image pairs.}
		(a) The evaluation results of IBE, SLM-BoF, Scale-Net\_COCO and Scale-Net\_ft\_PT in NDFF and PW.
		(b) The evaluation results of  SLM-BoF, Scale-Net\_COCO and Scale-Net\_ft\_PT in SC and SPS.
	}
	\label{fig:avg_diff_dist}
\end{figure}

We resort to Scale-MegaDepth dataset \cite{rau2020predicting} and Scale-IMC-PT dataset to compare the scale ratio estimation accuracy of the image box embedding model (IBE) \cite{rau2020predicting}, SLM-BoF \cite{zhou2017progressive} and our Scale-Net. 
The area pictured in PW of Scale-IMC-PT dataset is a part of that in Big Ben of Scale-MegaDepth dataset.
The area pictured in NDFF of Scale-IMC-PT dataset is a part of that in Notre-Dame of Scale-MegaDepth dataset. 
But PW and NDFF contain more image pairs with large scale differences than Big Ben and Notre-Dame.
Because IBE is not able to generalize across scenes, it is not evaluated in SC and SPS. To reimplement the SLM-BoF proposed by Zhou \etal, the training images of Scale-IMC-PT dataset are used to build a vocabulary tree. 8k SIFT features are extracted from every training image when building the vocabulary tree. When using SLM-BoF to estimate image scale ratios, we still extract 8k SIFT features from every image. 
We adopt the average $ L_1 $ discrepancy between the predicted and ground truth scale ratios as the evaluation metric here. Given ground truth scale ratios $ \left\{s_i\right\}_{i=1}^{N} $ and predicted scale ratios $ \left\{\hat{s}_{i}\right\}_{i=1}^{N} $, the average $ L_1 $ discrepancy, $ E $, is shown as below:

\begin{equation}\label{eq_srError}
E = \frac{1}{N}\sum_{i=1}^{N} \lvert \log_2{s_i} - \log_2{\hat{s}_i} \rvert
\end{equation}
The average $ L_1 $ discrepancies of IBE, SLM-BoF and Scale-Net are shown in Table \ref{tab_scaleRatioEst}. Scale-Net\_COCO is trained with Scale-COCO dataset. We further finetune Scale-Net\_COCO on the training set of Scale-IMC-PT dataset for only 1 epoch and obtain Scale-Net\_ft\_PT.
The curves of the average $ L_1 $ discrepancy over scale differences of image pairs are shown in Figure \ref{fig:avg_diff_dist}.
The meaning of a point $ \left( x\sim(x+1), y \right) $ on a curve is illustrated as follows: when only those image pairs whose ground truth scale ratios belong to $ \left(2^{-(x+1)}, 2^{-x}\right)\cup\left(2^x, 2^{x+1}\right) $ are taken into account, the average $ L_1 $ discrepancy is \textit{y}.

\begin{table}[htbp]
	\caption{Average L1 Discrepancies of IBE \cite{rau2020predicting}, SLM-BoF \cite{zhou2017progressive} and Our Scale-Net.}  
	\begin{center}
		\resizebox{\linewidth}{!}
		{
			\renewcommand{\arraystretch}{1.25}
			\begin{tabular}{l|cccc|cccc}
				\hline
				                                            &             \multicolumn{4}{c|}{Scale-MegaDepth}              &               \multicolumn{4}{c}{Scale-IMC-PT}                \\ \cline{2-9}
				\diagbox{\textbf{Methods}}{\textbf{Scenes}} &    BB    &  ND   &    Ve     &   Fl    &     NDFF      &      PW       &      SC       &      SPS      \\ \hline
				SLM-BoF                                     &     1.53      &     1.56      &     1.79      &     1.36      &     3.04      &     2.34      &     2.71      &     2.69      \\
				IBE                                         &     1.51      &     \textbf{1.23}      &     \textbf{1.43}      &     1.28      & 4.64 & 5.25 &       /       &       /       \\
				Scale-Net\_COCO                             &     1.52      &     1.42      &     1.72      &     1.39      & 2.24 & 1.71 & 1.58 & 1.83 \\
				Scale-Net\_ft\_PT                           & \textbf{1.32} & 1.26 & 1.51 & \textbf{1.06} & \textbf{1.08} & \textbf{1.12} & \textbf{0.74} & \textbf{0.86} \\ 
				\hline
			\end{tabular}
		}
	\end{center}
	\label{tab_scaleRatioEst}
\end{table}

\begin{figure*}[htbp]
	\centering
	\includegraphics[width=0.9\linewidth]{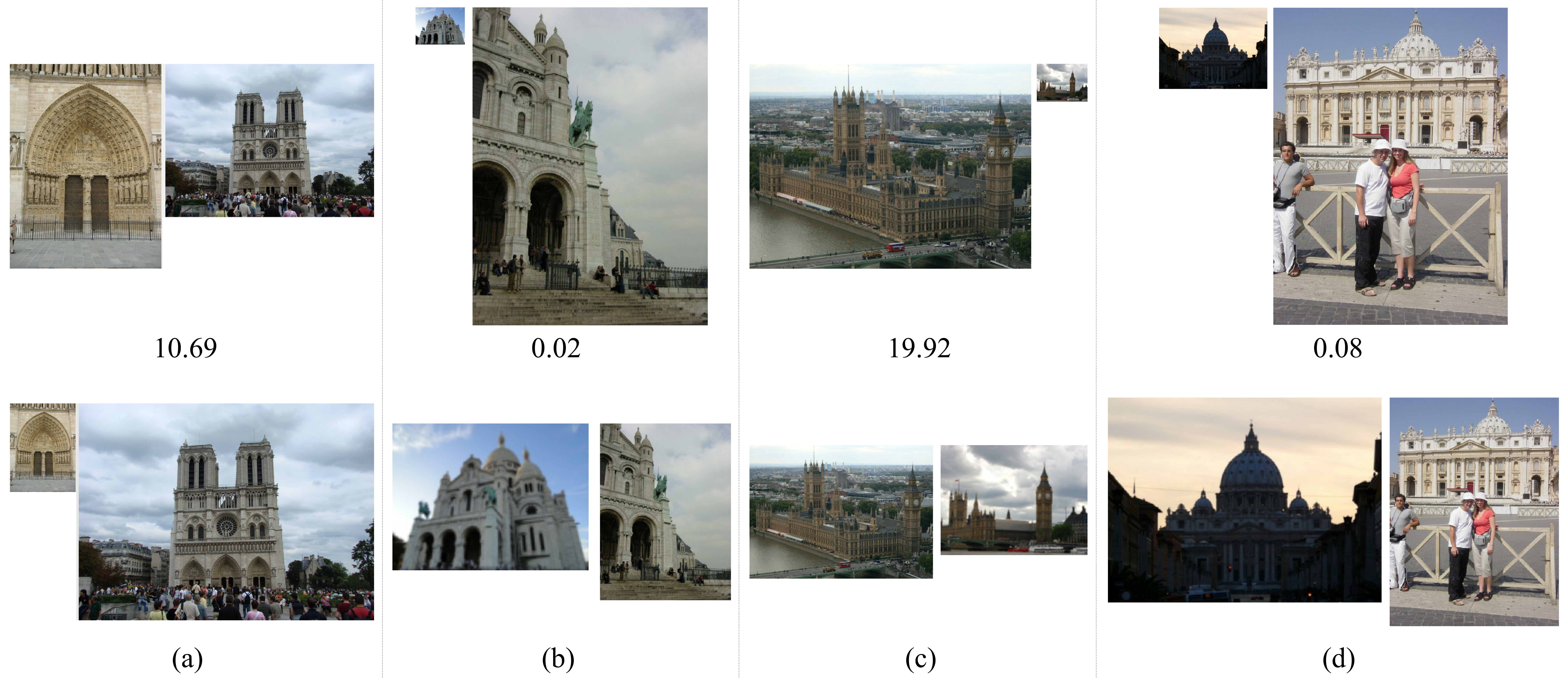}
	\caption{\textbf{Qualitative results of scale ratio estimation of Scale-Net on the test set of SCALE-IMC-PT dataset.}
	For the first row, we show four image pairs with considerable scale differences. In the first row, the number below each image pair is the scale ratio estimated by our Scale-Net. Each image pair is resized according to the estimated scale ratio. Resized image pairs are displayed in the second row.
	\textbf{Best viewed in color.}
}
	\label{fig:scaleest}
\end{figure*}

As shown in Table \ref{tab_scaleRatioEst}, our Scale-Net\_ft\_PT is comparable with IBE on Scale-MegaDepth dataset. 
Note that an IBE model can only apply to the single scene used for training it. Therefore, in this experiment, there are four IBE models trained with the four scenes of Scale-MegaDepth dataset respectively.
By contrast, our Scale-Net\_ft\_PT is not trained with Scale-MegaDepth dataset. Thus, compared with IBE, Scale-Net has much better generalization ability.
On Scale-IMC-PT dataset, even Scale-Net\_COCO is able to achieve higher accuracy than IBE and SLM-BoF, which further demonstrates the good generalization ability of Scale-Net.
As shown in Figure \ref{fig:avg_diff_dist}, compared with IBE and SLM-BoF, Scale-Net\_ft\_PT has much lower estimation error when facing large scale differences.
Several qualitative results of our Scale-Net\_ft\_PT are shown in Figure \ref{fig:scaleest}. 
The image pairs in the first row are resized according to the scale ratios estimated by our Scale-Net\_ft\_PT and displayed in the second row.
As shown in Figure \ref{fig:scaleest}, in every image pair, visual overlaps in two images take up approximately the same number of pixels, which means that the scale ratios estimated by Scale-Net\_ft\_PT are accurate.
And our Scale-Net\_ft\_PT is robust to viewpoint changes (Figure \ref{fig:scaleest}(b), (c)) and illumination changes (Figure \ref{fig:scaleest}(d)).

\subsection{Image Matching}
In this section, we utilize ES-HP dataset \cite{liu2019gift} to confirm that our SDAIM is able to boost the image matching performance of local features under extreme scale changes. 
We select SIFT \cite{lowe2004distinctive}, ASLFeat \cite{luo2020aslfeat} and GIFT-SP \cite{liu2019gift} as baselines. 
SIFT is a handcrafted feature detection algorithm whose robustness to scale changes is widely recognized. ASLFeat is a learning-based detect-and-describe feature extraction method which performs well on several broadly-used benchmarks \cite{balntas2017hpatches, 9229078, schonberger2017comparative}. GIFT-SP is a learning-based detect-then-describe feature extraction method, which uses Superpoint as the keypoint detector and uses GIFT as the descriptor and is robust to scale changes. 

\textbf{Evaluation Protocols.} 
Because IBE is not able to generalize across scenes, only SLM-BoF and our Scale-Net\_COCO are used to estimate scale ratios of image pairs for SDAIM. 
We assess the performance of the following two combinations: (SLM-BoF, SDAIM) and (Scale-Net\_COCO, SDAIM).
Due to the large distinction between ES-HP dataset and Scale-IMC-PT dataset, we build a new vocabulary tree for SLM-BoF using the training images of MS-COCO 2014 dataset. 8k SIFT features are extracted from every image during building the vocabulary tree. When using SLM-BoF to estimate image scale ratios, we still extract 8k SIFT features from every image.
Following GIFT \cite{liu2019gift}, we use Percentage of Correctly Matched Keypoints (PCK) to quantify the performance for correspondence estimation. 
PCK is defined as the ratio between the number of correct matches and the total number of interest points. 
Mutual nearest neighbour matching and ratio test are used to match local features. 
The evaluation results are shown in Table \ref{tab_esHP}.

\begin{table}[htbp]
	\caption{PCK of Local Features with or Without the Help of SDAIM on ES-HP Dataset.}  
	
	\begin{center}
		\resizebox{\linewidth}{!}
		{
			\renewcommand{\arraystretch}{1.25}
			\begin{tabular}{l|ccc}
				\hline
				&     Overall      &      Illumination       &      Viewpoint      \\ 
				\hline
				SIFT  &     15.40     &     12.80     &     17.90       \\
				SIFT + (SLM-BoF, SDAIM)&     15.72      &     12.84      &     18.52    \\
				SIFT + (Scale-Net\_COCO, SDAIM)&     \textbf{20.53}      &     \textbf{16.88}      &     \textbf{24.08}    \\
				
				\hline
				
				GIFT-SP &    27.28      &    25.63      &     28.89    \\
				GIFT-SP + (SLM-BoF, SDAIM)&     29.91      &     27.07     &    32.68    \\
				GIFT-SP + (Scale-Net\_COCO, SDAIM)&    \textbf{35.81}     &     \textbf{34.21}      &    \textbf{37.38}    \\
				
				\hline
				
				ASLFeat &     25.62      &     24.61      &    26.61     \\
				ASLFeat + (SLM-BoF, SDAIM)&     29.74      &     27.69      &    31.74   \\
				ASLFeat + (Scale-Net\_COCO, SDAIM)&     \textbf{36.31}      &     \textbf{35.85}      &     \textbf{36.75}   \\
				\hline		
			\end{tabular}
		}
	\end{center}
	\label{tab_esHP}
\end{table}

As shown in Table \ref{tab_esHP}, the results of SIFT, GIFT-SP and ASLFeat are all improved by SDAIM. And SDAIM with Scale-Net\_COCO is able to achieve much better results than SDAIM with SLM-BoF. The reason is that the image scale ratio accuracy of our Scale-Net is higher than that of SLM-BoF. Thus, SDAIM with Scale-Net can sample more related scale levels than SDAIM with SLM-BoF. In that way, with the aid of SDAIM and Scale-Net, more local features are extracted from related scale levels. More inlier correspondences can be established.
Table \ref{tab_esHP} also confirms that our Scale-Net is robust to viewpoint and illumination changes.

\begin{figure*}[htbp]
	\centering
	\includegraphics[width=0.9\linewidth]{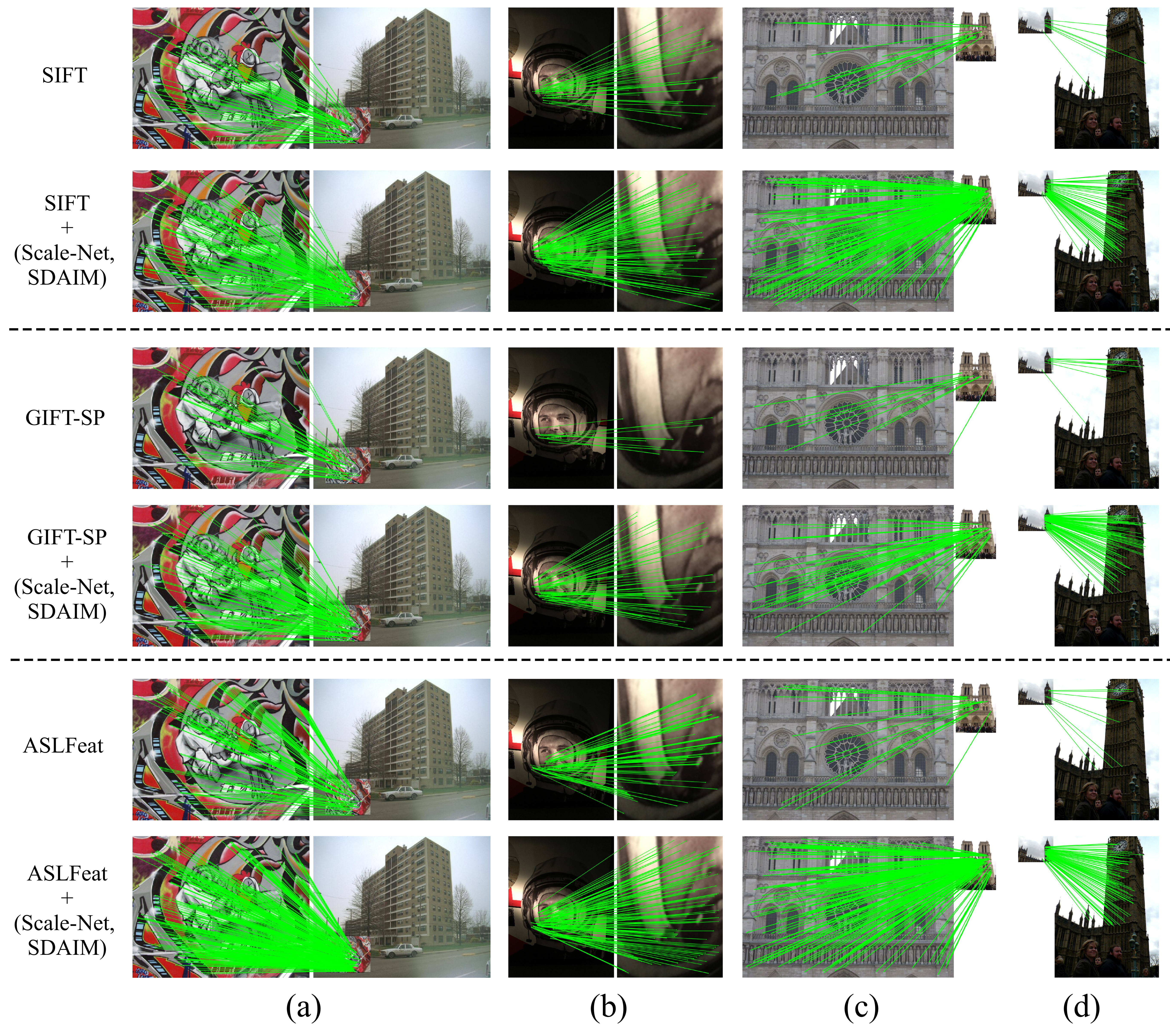}
	\caption{
		\textbf{Visualization of the estimated correspondences of SIFT, GIFT-SP (SuperPoint + GIFT) and ASLFeat, with or without the help of (Scale-Net, SDAIM).}
		(a) and (b) belong to ES-HP dataset. Scale-Net\_COCO is used to estimate their scale ratios. Only those correspondences conforming to the ground truth homography matrices are drawn.
		(c) and (d) belong to Scale-IMC-PT dataset. Scale-Net\_ft\_PT is used to estimate their scale ratios. Only those correspondences conforming to the ground truth epipolar geometry are drawn.
	}
	\label{fig:matchfeatures}
\end{figure*}

\subsection{Relative Pose Estimation}
\label{sec_features}

In this section, we demonstrate that SDAIM is able to improve the relative pose estimation precision of local features under large image scale changes. We evaluate the performance of SDAIM on the test set of Scale-IMC-PT dataset, which consists of 39778 image pairs whose ground truth relative camera poses are known. 
IBE, SLM-BoF and our Scale-Net are used to estimate scale ratios for SDAIM. We assess the performance of all the following combinations: (IBE, SDAIM), (SLM-BoF, SDAIM), (Scale-Net\_COCO, SDAIM), (Scale-Net\_ft\_PT, SDAIM).
In this section, SLM-BoF uses the vocabulary tree built in Section \ref{sec_sre}.
Because IBE is not able to generalize across scenes, it is not evaluated in SC and SPS.

\begin{figure}[htbp]
	\centering
	\includegraphics[width=0.95\linewidth]{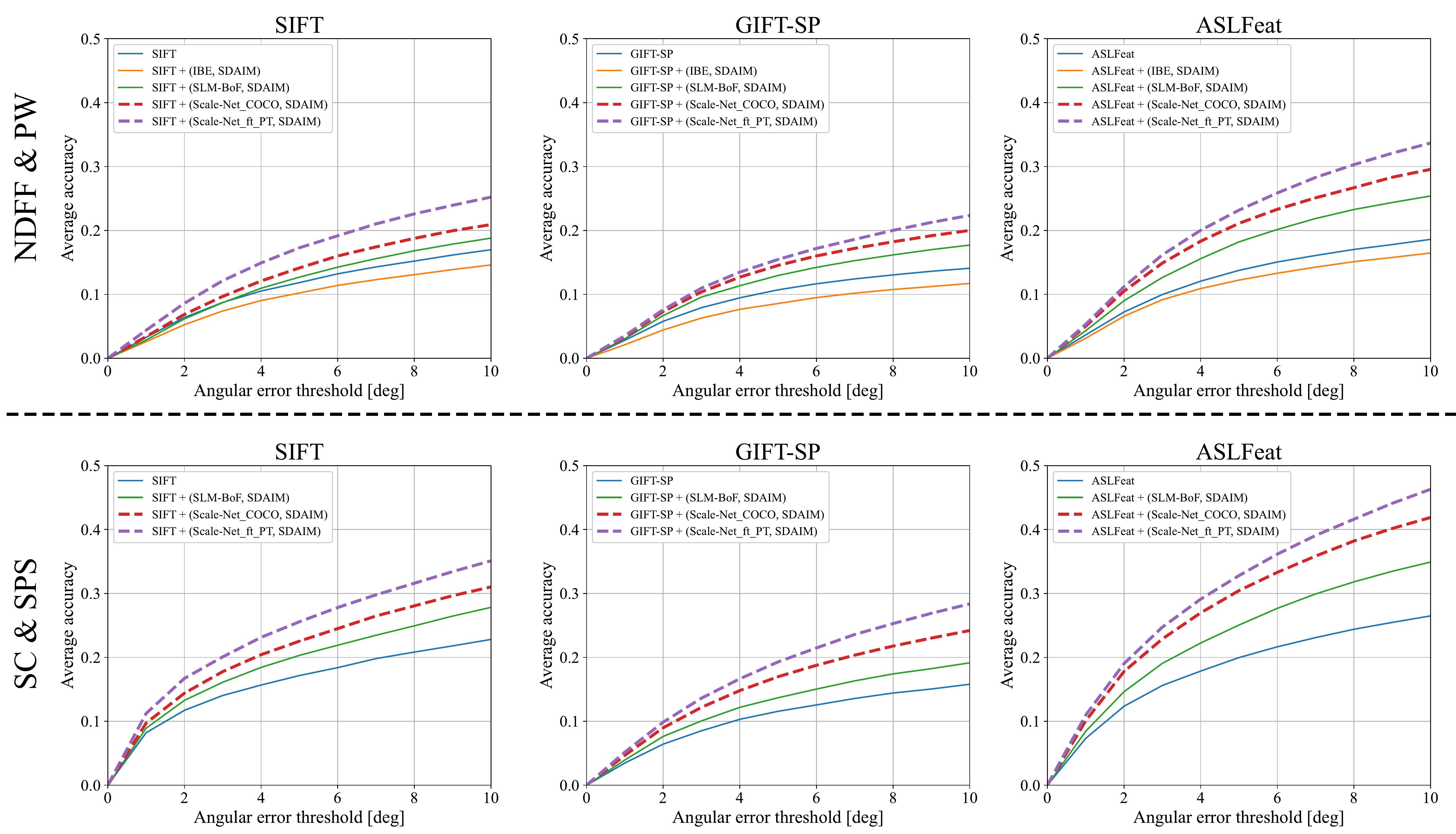}
	\caption{\textbf{The average accuracy curves of the estimated relative camera poses of local features.}
		In the first row, the average accuracy curves in NDFF and PW are shown.
		In the second row, the average accuracy curves in SC and SPS are shown.
		Because existing IBE models are not able to generalize to SC and SPS, (IBE, SDAIM) is not evaluated in SC and SPS.
		\textbf{Best viewed in color.}
	}
	\label{fig:aa_features}
\end{figure}

\textbf{Evaluation Protocols.} 
We select SIFT \cite{lowe2004distinctive}, ASLFeat \cite{luo2020aslfeat} and GIFT-SP \cite{liu2019gift} as baselines. 
2k feature points are extracted from every image. The feature matching procedure contains the following steps: mutual nearest neighbour matching (MNN), ratio test and RANSAC verification.
The RANSAC algorithm proposed by Fischler and  Bolles \cite{fischler1981random} is used. The reprojection threshold is 1.5 pixels. The confidence threshold is 0.999. The maximum number of iterations for random sample selection is 100000. 
Following the CVPR Image Matching Challenge \cite{jin2020image}, we compute the angular differences between the estimated and ground truth translation vectors, and between the estimated and ground truth rotation vectors, and take the largest of the two as the Final Pose Error (FPE). 
We draw average accuracy curves, as shown in Figure \ref{fig:aa_features}. A point $(x, y)$ on an average accuracy curve means that the ratio between the number of image pairs whose estimated relative poses are accurate within an error tolerance $x$, and the total number of test image pairs, is $y$.
In this experiment, the maximum error threshold is $10^{\circ}$.
We compute the mean average accuracy (mAA), the area under the average accuracy curve. The mAA($10^{\circ}$) of each method is shown in Table \ref{tab_mAA}.
The larger the mAA, the better the performance.

Both Table \ref{tab_mAA} and Figure \ref{fig:aa_features} show that our (Scale-Net\_COCO, SDAIM) and our (Scale-Net\_ft\_PT, SDAIM) are able to raise the relative pose estimation precision of all the three representative local features who have laid much stress on scale invariance. Both (Scale-Net\_COCO, SDAIM) and our (Scale-Net\_ft\_PT, SDAIM) perform better than (SLM-BoF).
(IBE, SDAIM) slightly worsen the relative pose estimation performance of local features. 
Combining Table \ref{tab_scaleRatioEst} and Table \ref{tab_mAA}, we can find that the performance of our SDAIM relies heavily on the image scale ratio estimation precision. The higher the scale ratio estimation precision, the better the performance of our SDAIM. And the low scale ratio estimation precision of IBE results in the bad performance of (IBE, SDAIM).
The reasons are as follows. Local feature matching plays a vital role in relative pose estimation task.
The scale ratio estimation method with higher precision enables our SDAIM to sample more related scale levels of image pairs. In that way, according to Section \ref{sec_SDAIM}, more inlier correspondences can be established, as shown in Figure \ref{fig:matchfeatures}(c) and (d).
However, the scale ratio estimation method with low precision makes SDAIM to sample less related scale levels.

\begin{figure}[htbp]
	\centering
	\includegraphics[width=0.95\linewidth]{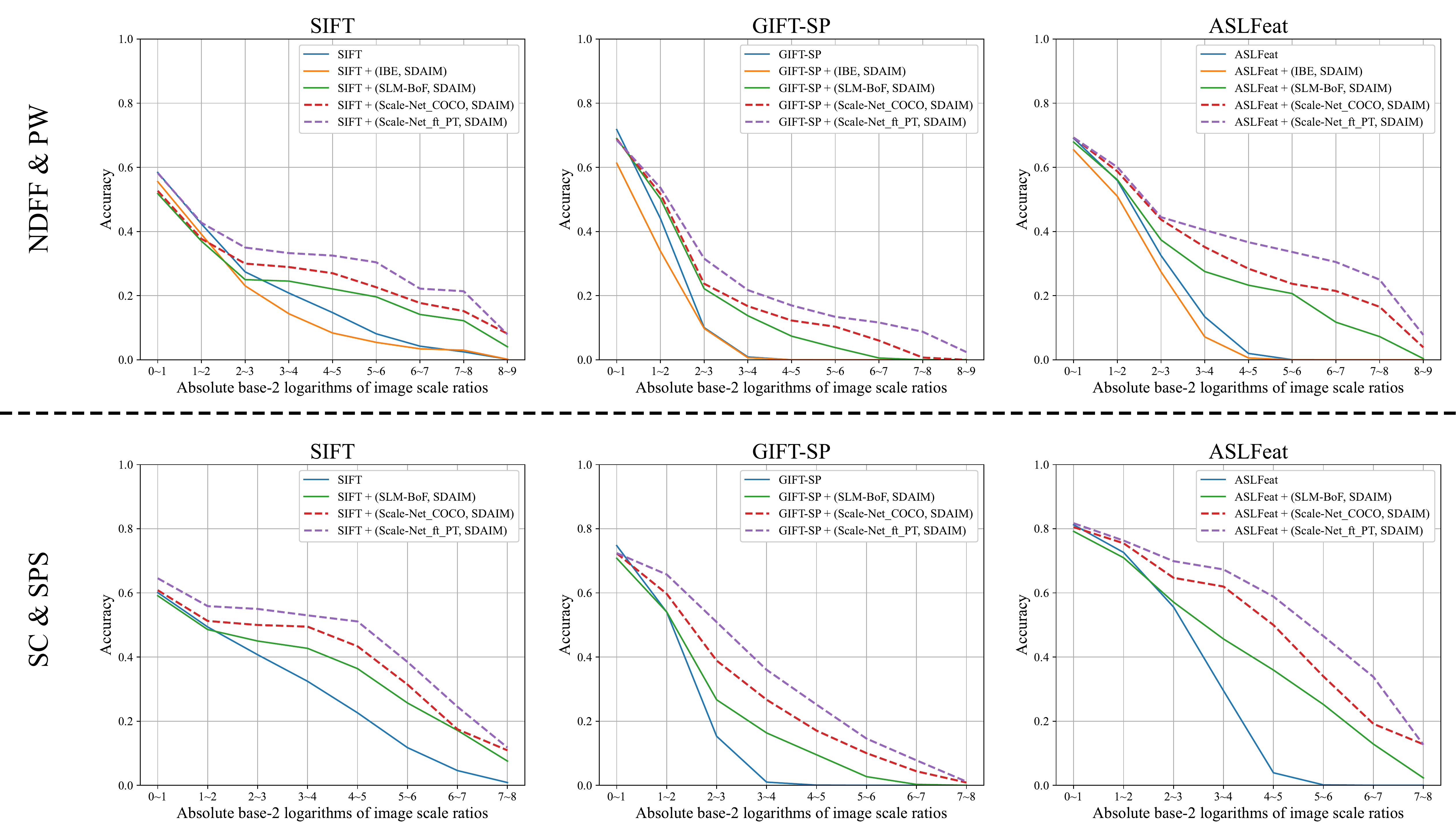}
	\caption{\textbf{The accuracy curves of the estimated relative camera poses of several local features.}
		If the FPE between the estimated pose and the ground truth pose is smaller than $20^{\circ}$, the estimated pose is regarded as accurate.
		In the first row, the accuracy curves in NDFF and PW are shown.
		In the second row, the accuracy curves in SC and SPS are shown. 
		Because existing IBE models are not able to generalize to SC and SPS, (IBE, SDAIM) is not evaluated in SC and SPS.
		\textbf{Best viewed in color.}
	}
	\label{fig:aa_features_dist}
\end{figure}

If the FPE between the estimated pose and the ground truth pose is smaller than $ 20^\circ $, the estimated pose is regarded as accurate. The curves of relative pose estimation accuracy over scale differences of image pairs are drawn in Figure \ref{fig:aa_features_dist}. 
According to Figure \ref{fig:aa_features_dist}, when absolute base-2 logarithms of image scale ratios are larger than 2, our (Scale-Net\_COCO, SDAIM) and (Scale-Net\_ft\_PT, SDAIM) are able to raise the relative pose estimation accuracy of all the above local features. The boost from SLM-BoF is not so obvious compared with our Scale-Net.

\begin{table}[htbp]
	\caption{Relative Pose Estimation Results of Several Local Features on the Test Set of SCALE-IMC-PT Dataset.}  
	
	\begin{center}
		\resizebox{\linewidth}{!}
		{
			\renewcommand{\arraystretch}{1.25}
			\begin{tabular}{l|cccc}
				\hline
				\diagbox{\textbf{Methods}}{\textbf{mAA}(10$^{\circ}$)}{\textbf{Scenes}} &     NDFF      &      PW       &      SC       &      SPS      \\ 
				\hline
				SIFT  &     0.099      &     0.078      &     0.223      &     0.081      \\
				SIFT + (IBE, SDAIM)                           &     0.101      &     0.075     &     /      &     /      \\
				SIFT + (SLM-BoF, SDAIM)&     0.129      &     0.091      &     0.261      &     0.098   \\
				SIFT + (Scale-Net\_COCO, SDAIM)&    0.147      &     0.098      &     0.290      &    0.109    \\
				SIFT + (Scale-Net\_ft\_PT, SDAIM)&     \textbf{0.164}      &     \textbf{0.134}      &     \textbf{0.322}      &     \textbf{0.131}  \\
				
				\hline
				
				GIFT-SP &     0.090      &     0.090      &     0.132      &     0.065     \\
				GIFT-SP + (IBE, SDAIM)                           &     0.077      &     0.069      &     /      &      /     \\
				GIFT-SP + (SLM-BoF, SDAIM)&     0.118      &     0.101      &     0.158      &     0.078   \\
				GIFT-SP + (Scale-Net\_COCO, SDAIM)&     0.135      &     0.111      &     0.201      &    0.092    \\
				GIFT-SP + (Scale-Net\_ft\_PT, SDAIM)&     \textbf{0.148}      &     \textbf{0.117}      &     \textbf{0.221}      &     \textbf{0.114}   \\
				
				\hline
				
				ASLFeat &     0.137      &     0.096      &     0.238     &   0.108   \\
				ASLFeat + (IBE, SDAIM)                           &     0.126      &     0.081      &     /      &      /     \\
				ASLFeat + (SLM-BoF, SDAIM)&     0.190      &      0.119     &     0.297      &    0.141    \\
				ASLFeat + (Scale-Net\_COCO, SDAIM)&     0.218      &     0.140      &     0.364      &    0.163    \\
				ASLFeat + (Scale-Net\_ft\_PT, SDAIM)&     \textbf{0.245}      &     \textbf{0.153}      &     \textbf{0.388}      &     \textbf{0.185}   \\	
				\hline			
			\end{tabular}
		}
	\end{center}
	\label{tab_mAA}
\end{table}

\subsection{Enhancements to Local Feature Matching Methods}

\begin{figure*}[htbp]
	\centering
	\includegraphics[width=0.9\linewidth]{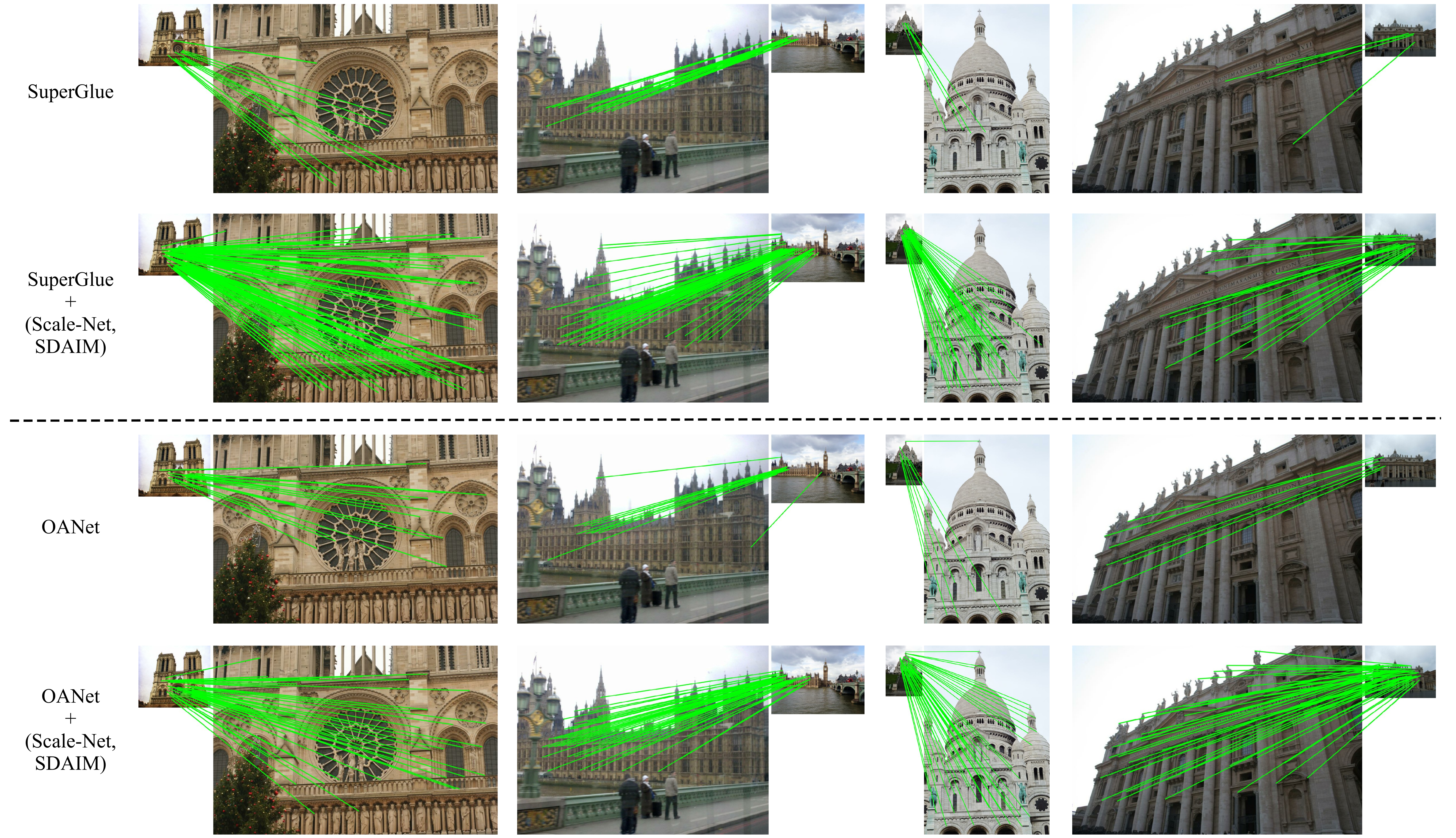}
	\caption{
		\textbf{Visualization of the estimated correspondences of SuperGlue and OANet, with or without the help of (Scale-Net, SDAIM).}
		Only those correspondences conforming to the ground truth epipolar geometry are drawn.
		SuperGlue is used to match SuperPoint feature points. OANet is used to match SIFT feature points.
	}
	\label{fig:matcherresize}
\end{figure*}

In previous experiments, only mutual nearest neighbor matching (MNN) and ratio test are used to match local features. 
In this experiment, we resort to the test set of Scale-IMC-PT dataset to confirm that our Scale-Net and SDAIM are able to greatly enhance relative pose estimation performance of various state-of-the-art local feature matching methods under large image scale changes.

\begin{figure}[htbp]
	\centering
	\includegraphics[width=0.95\linewidth]{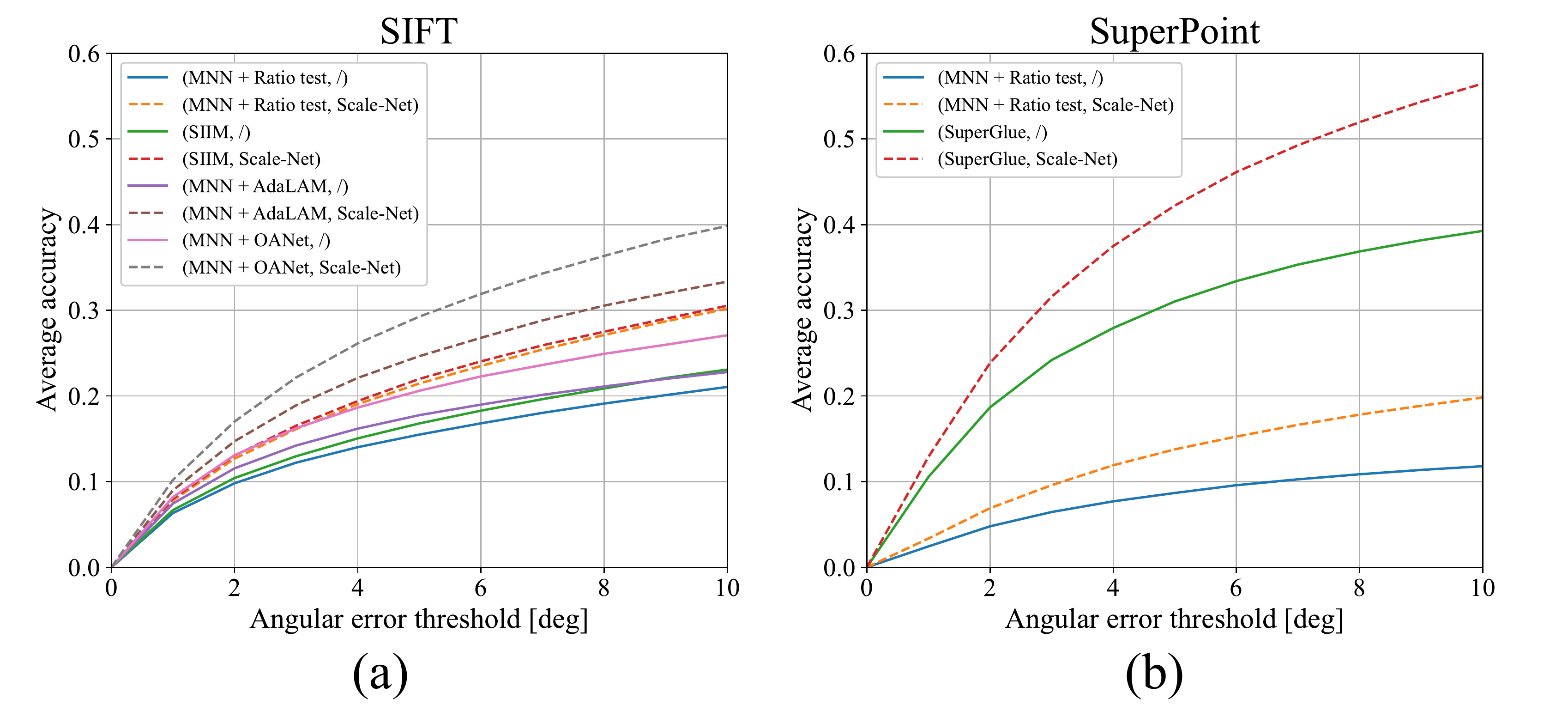}
	\caption{\textbf{The average accuracy curves of the estimated relative camera poses of various local feature matching methods.}
		(a) The adopted local feature is SIFT.
		(b) The adopted local feature is SuperPoint.
		\textbf{Best viewed in color.}
	}
	\label{fig:aa_matchers}
\end{figure}

\textbf{Evaluation Protocols.} We select SIIM \cite{zhou2017progressive}, AdaLAM \cite{cavalli2020handcrafted}, OANet \cite{9310246} and SuperGlue \cite{sarlin2020superglue} as baselines.
To the best of our knowledge, the scale-invariant image matching (SIIM) proposed by Zhou \etal \cite{zhou2017progressive} is the only existing local feature matching method focusing on image scale changes. AdaLAM \cite{cavalli2020handcrafted} is the existing best handcrafted outlier rejection method. OANet \cite{9310246} is the existing best learned outlier rejection method.
SuperGlue \cite{sarlin2020superglue} is the existing best learned local feature matching method. 
The adopted local features are SIFT \cite{lowe2004distinctive} and SuperPoint \cite{detone2018superpoint}. 2k feature points are extracted from every image.
We draw average accuracy curves in Figure \ref{fig:aa_matchers} and report mAA($10^{\circ}$) in Table \ref{tab_matcher}. 
In Table \ref{tab_matcher}, / means that SDAIM is not used. In another word, / means that image pairs are not resized before local feature extraction.
For convenience, we use a tuple, (\textit{m}, Scale-Net), to denote the matching method, \textit{m}, assisted by SDAIM and Scale-Net\_ft\_PT. (\textit{m}, /) means that \textit{m} is the adopted matching method and that image pairs are not resized before local feature extraction.

As shown in Figure \ref{fig:aa_matchers}(a) and Table \ref{tab_matcher}, compared with (SIIM, /), (MNN + Ratio test, Scale-Net) achieves better results. 
By comparing the results of (SIIM, /) and (SIIM, Scale-Net), we can find that SDAIM and Scale-Net are able to boost the performance of SIIM.
And SDAIM and Scale-Net can also greatly improve the performance of AdaLAM and OANet. As shown in Figure \ref{fig:aa_matchers}(b) and Table \ref{tab_matcher}, our SDAIM and Scale-Net can still drastically enhance the performance of SuperGlue.
Qualitative results of the remarkable boost from (Scale-Net, SDAIM) are shown in Figure \ref{fig:matcherresize}. Our Scale-Net, and SDAIM greatly raise the inlier correspondence numbers of SuperGlue and OANet.

\begin{figure}[htbp]
	\centering
	\includegraphics[width=0.95\linewidth]{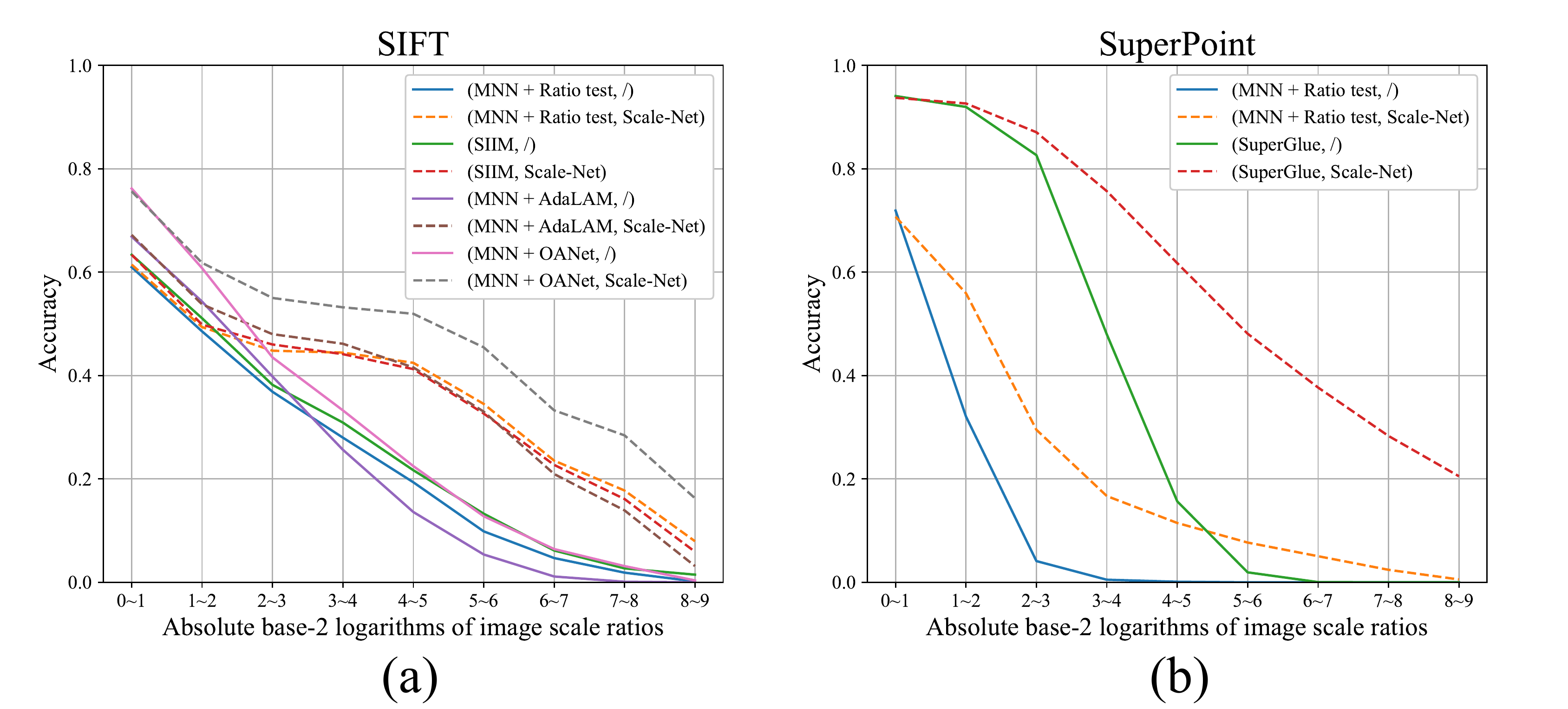}
	\caption{\textbf{The accuracy curves of the estimated relative camera poses of various local feature matching methods.}
		If the FPE between the estimated pose and the ground truth pose is smaller than $20^{\circ}$, the estimated pose is regarded as accurate.
		(a) The adopted local feature is SIFT.
		(b) The adopted local feature is SuperPoint.
		\textbf{Best viewed in color.}
	}
	\label{fig:aa_dist}
\end{figure}

Following Section \ref{sec_features}, we also draw the curves of relative pose estimation accuracy over scale differences of image pairs, as shown in Figure \ref{fig:aa_dist}.
When absolute base-2 logarithms of image scale ratios are larger than 2, our proposed SDAIM and Scale-Net are able to drastically raise the relative pose estimation accuracy of all the above local feature matching methods.

\begin{table}[htbp]
	\caption{Relative Pose Estimation Results of Several Representative Local Feature Matching Methods on the Test Set of Scale-IMC-PT Dataset.}  
	
	\begin{center}
		\resizebox{\linewidth}{!}
		{
			\renewcommand{\arraystretch}{1.25}
			\begin{tabular}{clccccc}
				\toprule[1pt]
				\multirow{2}*{\shortstack{Local \\ Features}}    &      \multirow{2}*{Matchers}           &       \multirow{2}*{SDAIM}   & \multicolumn{4}{c}{Scenes} \\
				\cline{4-7}
				                                                 &                                       &                              & NDFF & PW & SC & SPS \\
				\midrule[0.5pt]
				\multirow{8}*{SIFT}                             &     MNN + Ratio test                  &     /                        & 0.099 & 0.078 & 0.223 & 0.131 \\
						
				                                             & SIIM                       &     /                        & 0.112 & 0.100 & 0.215 & 0.091 \\
				                                             
				                                             & MNN + Ratio test                      &     Scale-Net                & 0.164 & 0.134 & 0.322 & 0.131 \\
				
%
%
				                                             & SIIM                                  &     Scale-Net                & 0.171 & 0.141 & 0.311 & 0.142 \\
				
				                                             & MNN + AdaLAM                                &     /                        & 0.131 & 0.120 & 0.250 & 0.112 \\
				
				                                             & MNN + AdaLAM                                &     Scale-Net                & 0.196 & 0.161 & 0.332 & 0.166 \\
				
				                                             & MNN + OANet                                 &     /                        & 0.160 & 0.153 & 0.261 & 0.141 \\
				
				                                             & MNN + OANet                                 &     Scale-Net                & \textbf{0.239} & \textbf{0.207} & \textbf{0.362} & \textbf{0.205} \\
				\hline
				\multirow{4}*{SuperPoint}                       & MNN + Ratio test                        &     /                        & 0.071 & 0.069 & 0.105 & 0.052 \\
				
				                                       & MNN + Ratio test                        &     Scale-Net                & 0.118 & 0.093 & 0.170 & 0.092 \\
				
				                                       & SuperGlue                             &     /                        & 0.232 & 0.251 & 0.367 & 0.201 \\
				
				                                       & SuperGlue                             &     Scale-Net                & \textbf{0.320} & \textbf{0.329} & \textbf{0.517} & \textbf{0.274} \\
				\bottomrule[1pt]
			\end{tabular}
		}
	\end{center}
	\label{tab_matcher}
\end{table}

\subsection{Ablation Study}
In this section, we confirm the effectiveness of our Covisibility-Attention-Reinforced Matching Module, abbreviated as CVARM.
We remove $ CAB_1 $ and $ CAB_2 $ in Figure \ref{fig:scale-net}(c) and obtain Scale-Net without CVARM, denoted by Scale-Net\_plain. 
And Scale-Net\_plain is trained in the same way as Scale-Net\_COCO.
Then, Scale-Net\_plain is evalutated on the test set of Scale-IMC-PT dataset and compared with Scale-Net\_COCO.
Scale-Net\_plain and Scale-Net\_COCO are used to estimate image scale ratios for SDAIM respectively. (Scale-Net\_plain, SDAIM) and (Scale-Net\_COCO, SDAIM) are used to boost the relative pose estimation performance of SuperGlue \cite{sarlin2020superglue}. The evaluation results of Scale-Net\_plain and Scale-Net\_COCO are shown in Table \ref{tab_ab}.
$ E $ denotes the scale ratio estimation error, as shown in Eq. (\ref{eq_srError}).

\begin{table}[htbp]
	\caption{Evaluation Results of Scale-Net with or Without CVARM on the Test Set of Scale-IMC-PT Dataset.} 
	
	\begin{center}
		\resizebox{\linewidth}{!}
		{
			\renewcommand{\arraystretch}{1.25}
			\begin{tabular}{c|c|c|c|c|c|c|c|c}
				\hline
				\multirow{2}*{Networks} &     \multicolumn{2}{c|}{NDFF}      &      \multicolumn{2}{c|}{PW}       &      \multicolumn{2}{c|}{SC}   &    \multicolumn{2}{c}{SPS}      \\ 
				\cline{2-9}
				                     & \textit{E}  & mAA($ 10^{\circ} $) & \textit{E} & mAA($ 10^{\circ} $) &  \textit{E}  & mAA($ 10^{\circ} $)     & \textit{E} & mAA($ 10^{\circ} $) \\
				\hline
				Scale-Net\_plain			 &     3.04    &   0.277     &     1.99    &      0.308         &   2.19  &  0.451           & 2.28     &           0.242             \\
				Scale-Net\_COCO           & \textbf{2.24} & \textbf{0.291} & \textbf{1.71} & \textbf{0.324} & \textbf{1.58} & \textbf{0.483} & \textbf{1.83} & \textbf{0.253}        \\
				\hline	
			\end{tabular}
		}
	\end{center}
	\label{tab_ab}
\end{table}

\begin{figure}[htbp]
	\centering
	\includegraphics[width=0.95\linewidth]{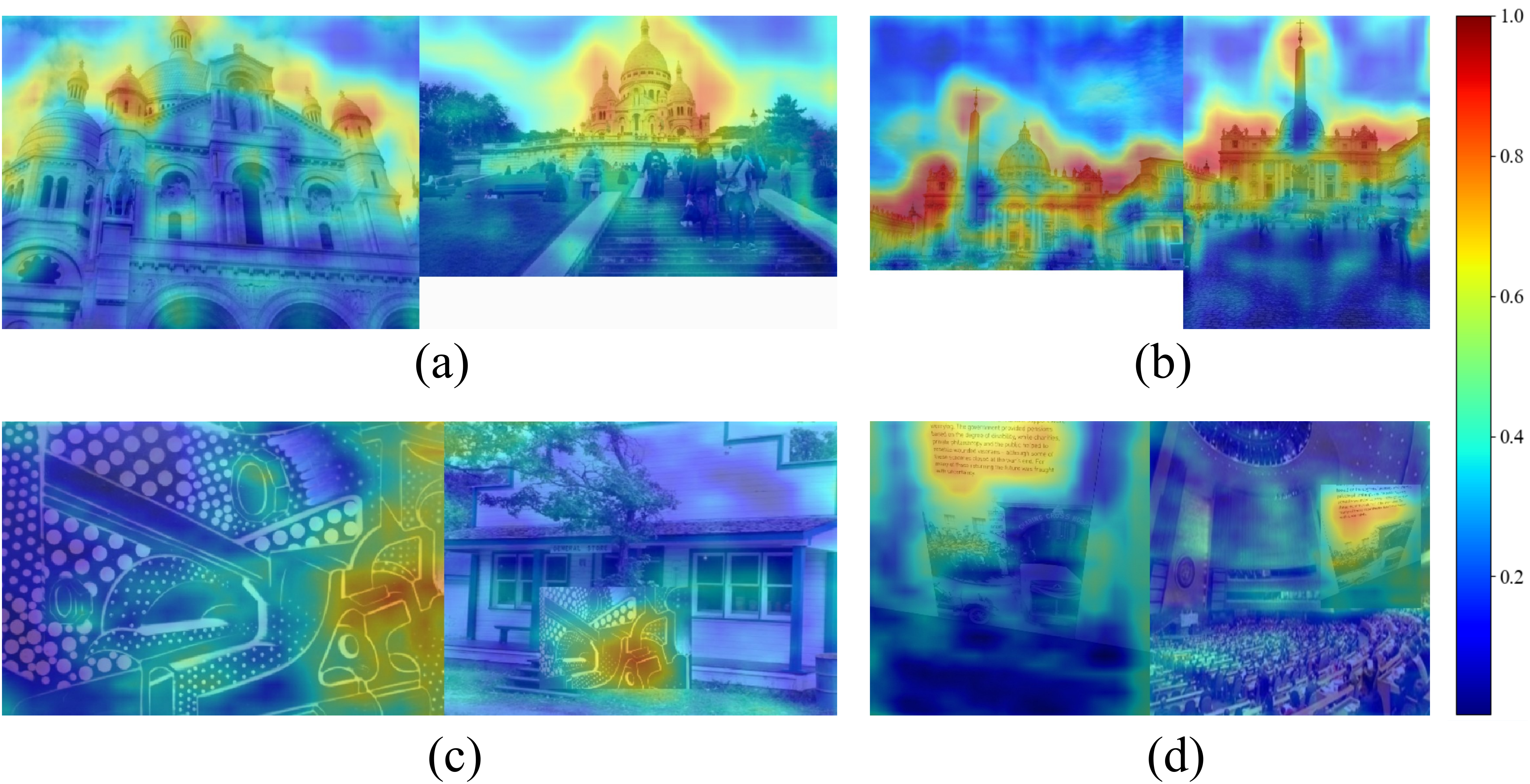}
	\caption{\textbf{Visualization of Covisibility Attention.} 
		There are four image pairs overlaid with covisibility attention maps.
		(a) and (b) are two image pairs in the test set of Scale-IMC-PT dataset.
		(c) and (d) are two image pairs in ES-HP dataset \cite{liu2019gift}.
		\textbf{Best viewed in color.}
	}
	\label{fig:attentionmaps}
\end{figure}

As shown in Table \ref{tab_ab}, Scale-Net\_COCO has higher scale ratio estimation accuracy than Scale-Net\_plain in the four scenes. And Scale-Net\_COCO achieves higher relative pose estimation accuracy than Scale-Net\_plain. These results demonstrate that the proposed CVARM is able to improve the scale ratio estimation accuracy of Scale-Net. Note that our CVARM only contains 50 parameters.
In Figure \ref{fig:attentionmaps}, four image pairs overlaid with covisibility attention maps, i.e., $ \textit{\textbf{M}}_1 $ and $ \textit{\textbf{M}}_2 $ in Figure \ref{fig:scale-net}(c), are displayed. Figure \ref{fig:attentionmaps} shows that the proposed CVARM attends to the visual overlaps within an image pair and suppresses the distraction from those regions visible in only one image indeed.

\section{Conclusion}

In this paper, we propose a scale-difference-aware image matching (SDAIM) approach that is able to greatly improve the performance of local features under large scale changes in images.
Given an image pair, firstly, it estimates the scale ratio of this image pair. Secondly, it resizes both images to drastically reduce the scale difference within the image pair before local feature extraction so that it can still establish sufficient inlier correspondences for downstream tasks when facing large scale changes.
We also propose a novel neural network, termed as Scale-Net, based on our proposed covisibility attention mechanism, to accurately estimate the scale ratios of image pairs for the proposed SDAIM. 
Rigorous evaluations on image matching and relative pose estimation tasks have demonstrated the remarkable performance of SDAIM, the good generalizaion ability and the high scale ratio estimation accuracy of Scale-Net.
In the future, we will use the Transformer architecture to rebuild Scale-Net in order to estimate image scale ratios more accurately.

{
\bibliographystyle{IEEEtran}
\bibliography{references}
}

\textbf{Supplementary Appendix}

We display our ICCV2021 submission record here to confirm that our idea is originated before March, 2021.

\begin{figure*}[b]
	\centering
	\includegraphics[width=0.95\linewidth]{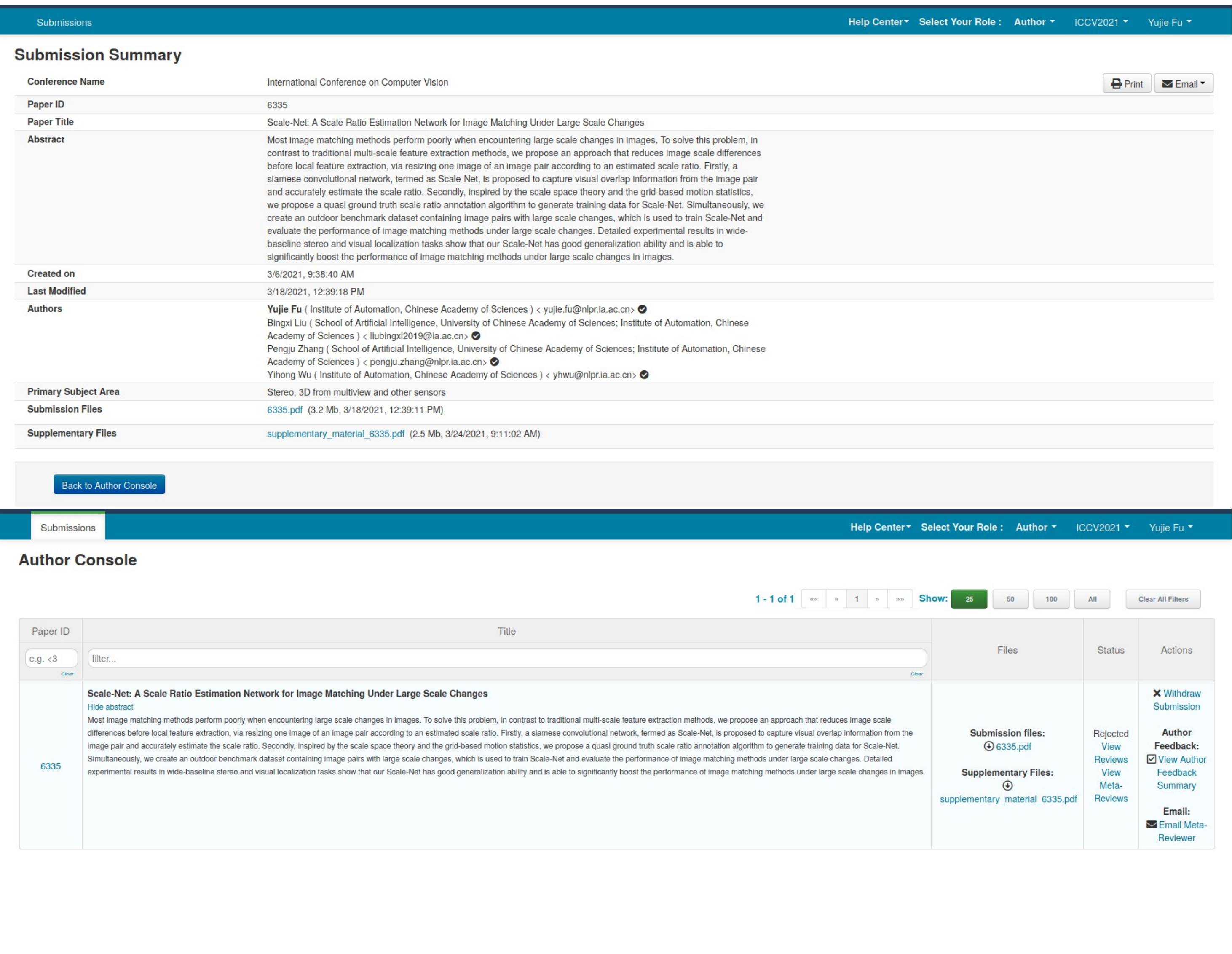}
	\label{fig:iccvsubmission0}
\end{figure*}

\vfill

\end{document}